\documentclass[accepted]{uai2022} 

\usepackage{hyperref}       
\usepackage{nameref} 
\usepackage{zref-xr}
\zxrsetup{toltxlabel} 
\zexternaldocument*{suehiro_124-supp}

\usepackage[american]{babel}

\usepackage{natbib} 
\usepackage{mathtools} 
\mathtoolsset{showonlyrefs=true}
\usepackage{booktabs} 
\usepackage{tikz} 

\usepackage{amssymb}

\usepackage[utf8]{inputenc} 
\usepackage[T1]{fontenc}    
\usepackage{url}            
\usepackage{booktabs}       
\usepackage{amsfonts}       
\usepackage{nicefrac}       
\usepackage{microtype}      
\usepackage{xcolor}         

\usepackage{times}
\usepackage{lscape}
\usepackage{booktabs}       
\usepackage{amsfonts}       
\usepackage{nicefrac}       
\usepackage{microtype}      
\usepackage{caption}
\usepackage{enumitem}
\usepackage{amsfonts}
\usepackage{amsmath}
\usepackage{braket}
\usepackage{boxedminipage}
\usepackage{epsf}
\usepackage{bm}
\usepackage{amsthm}
\usepackage{xspace}
\usepackage{wrapfig}
\usepackage{algorithm,algpseudocode}
\usepackage{multirow}
\algnewcommand{\Inputs}[1]{%
  \State \textbf{Inputs:}
  \Statex \hspace*{\algorithmicindent}\parbox[t]{.8\linewidth}{\raggedright #1}
}
\algnewcommand{\Initialize}[1]{%
  \State \textbf{Initialize:}
  \Statex \hspace*{\algorithmicindent}\parbox[t]{.8\linewidth}{\raggedright #1}
}
\def\indot<#1>{\langle #1 \rangle}

\newtheorem{defi}{Definition}
\newtheorem{theo}{Theorem}
\newtheorem{prop}[theo]{Proposition}
\newtheorem{coro}[theo]{Corollary}

\newtheorem{lemm}{Lemma}

\def\OMIT#1{}

\newcounter{nombre}
\renewcommand{\thenombre}{\arabic{nombre}}
\newcommand{\bx}{{\mathbf x}}
\newcommand{\by}{{\mathbf y}}
\newcommand{\bzero}{{\mathbf 0}}
\newcommand{\bmu}{\boldsymbol{\mu}}

\newcommand{\sign}{\mathrm{sign}}
 
\newcommand{\Real}{\mathbb{R}} 
\newcommand{\Hilbert}{\mathbb{H}} 

\newcommand{\F}{\mathrm{False}}
\newcommand{\T}{\mathrm{True}}
\newcommand{\calL}{\mathcal{L}}

\newcommand{\calG}{\mathcal{G}}

\newcommand{\calY}{\mathcal{Y}}

\newcommand{\vecsigma}{\boldsymbol{\sigma}}

\newcommand{\hata}{\hat{a}}

\newcommand{\Expo}{\mathop{\mathbb{E}}\limits}

\newcommand{\calX}{\mathcal{X}}
\newcommand{\calZ}{\mathcal{Z}}
\newcommand{\calV}{\mathcal{V}}
\newcommand{\calH}{\mathcal{H}}

\newcommand{\calD}{\mathcal{D}}
\newcommand{\Rdm}{\mathfrak{R}}

\newcommand{\calhatH}{\widehat{\mathcal{H}}}
\newcommand{\calhatG}{\widehat{\mathcal{G}}}

\newcommand{\RMC}{R^{\mathrm{MC}}}
\newcommand{\RLC}{R^{\mathrm{LC}}}

\newcommand{\emRLC}{\widehat{R}^{\mathrm{LC}}}

\newcommand{\emR}{\widehat{R}}

\newcommand{\GMIL}{MIL\xspace}

\title{Simplified and Unified Analysis of Various Learning Problems \\by Reduction to Multiple-Instance Learning}

\author[1, 2]{Daiki Suehiro}

\affil[1]{Kyushu University, Department of Advanced Information Technology, 
            744 Motooka, 
            Fukuoka, Japan}
\affil[2]{RIKEN, Center for Advanced Intelligence Project, 
            Nihonbashi 1-chome Mitsui Building, 15th floor,1-4-1 Nihonbashi, Chuo-ku, 
            Tokyo, Japan}

\author[3]{Eiji Takimoto}
\affil[3]{Kyushu University, Department of Informatics,
            744 Motooka, 
            Fukuoka,
            Japan}

\begin{document}
\maketitle

\begin{abstract}
In statistical learning, many problem formulations have been proposed so far,
such as multi-class learning, complementarily labeled learning, multi-label learning, multi-task learning, which provide theoretical models for various real-world tasks. 
Although they have been extensively studied, the relationship among them
has not been fully investigated.
In this work, we focus on a particular problem formulation called
Multiple-Instance Learning (MIL), and show that various learning problems
including all the problems mentioned above with some of new problems can be
reduced to MIL with theoretically guaranteed generalization bounds,
where the reductions are established under a new reduction scheme
we provide as a by-product. 
The results imply that the MIL-reduction gives a simplified and
unified framework for designing and analyzing algorithms for various
learning problems.
Moreover, we show that the MIL-reduction framework can be kernelized.
\end{abstract}

\section{Introduction}
\label{sec:intro}
In this study, we explore how a large class of learning problems can be reduced to the Multiple-Instance Learning (MIL) problem.
This is strongly motivated by the results of~\citep{Sabato:2012:MLA} and~\citep{suehiro2020multiple}.
\citet{suehiro2020multiple} showed that some local-feature-based learning
problems can be reduced to a MIL problem, which gave us an insight that MIL would
have a high capability of representing various learning problems.
Indeed, the reduced problem is too specific whereas \citet{Sabato:2012:MLA} proposed
a much more general formulation of MIL,
and thus we believe that a wider class of learning problems can be reduced to MIL.
\par
We provide a MIL-reduction scheme and reveal that various learning problems,
such as multi-class learning, complementarily labeled learning, multi-label learning, and multi-task learning, can be reduced to MIL.
By the reduction, we immediately derive generalization bounds from~\citep{Sabato:2012:MLA}, as well as learning algorithms. 
That is, our reduction scheme greatly \emph{simplifies} the analyses of generalization bounds as compared with the analyses in the previous works~\citep[e.g.,][]{Lei19,ishida2017learning,pmlr-v32-yu14,pontil2013excess}.
Some of the obtained generalization bounds are competitive or incomparable to the existing results. 
In particular, for multi-label learning, we derive an improved generalization bound, and for complementarily labeled learning, we derive a novel learning algorithm, which is the first polynomial-time algorithm in a certain setting.
Moreover, we propose three new learning problems, \emph{multi-label learning with perfectionistic loss}, \emph{top-1 ranking learning} and \emph{top-1 ranking learning with negative feedback},
and we demonstrate that they can be reduced to MIL as well.
The results imply that our MIL-reduction gives a \emph{unified scheme} for designing and analyzing algorithms for various learning problems.
\par
To provide the MIL-reduction scheme, we propose a general reduction scheme among learning problems.
Our scheme has two remarkable features as described below.
First, our reduction transforms every instance-label pair $(x,y)$
in the given sample of the original learning problem to an instance-label pair $(x',y')$ to form a sample of the reduced learning problem. 
In contrast, standard reduction schemes employ an instance
transformation and an label transformation separately, to construct
$x'$ from $x$ and $y'$ from $y$, respectively.
Therefore, our scheme enables us to design reduction algorithms
among a wider class of learning problems, e.g., learning-to-rank
to classification, and supervised learning to weakly supervised learning.
Second, our reduction scheme ensures that
the Empirical Risk Minimization (ERM) of the reduced problem implies the ERM of the original one, while the empirical Rademacher complexity
of the hypothesis (composed with loss function) classes are preserved through the reduction.
This means that we can employ an existing ERM algorithm for the
reduced problem to obtain an ERM algorithm for the original problem
with a theoretical guaranteed generalization bound, which is
immediately derived from a known generalization bound for the reduced problem.
We also show that the MIL-reduction scheme can be kernelized. 
\par
The main contributions are summarized as follows:
\begin{itemize}
\item We propose a general reduction scheme based on the ERM, which allows us to
derive a generalization risk bound of the original problem immediately.
\item We demonstrate that several learning problems, from traditional to new problems, can be reduced to MIL. The results imply that our MIL-reduction gives a simplified and unified scheme for the analyses for various learning problems.
\item We obtain novel theoretical results for some learning problems.
\item We show that the MIL-reduction scheme can be kernelized.
\end{itemize}
Several proofs are shown in supplementary materials.

\section{Preliminaries}
\label{sec:prelim}
For an integer $u$, $[u]$ denotes the set $\{1,\ldots, u\}$.
$I(\mathrm{e})$ denotes the indicator function of the event $\mathrm{e}$,
that is, $I(\mathrm{e})=1$ if $\mathrm{e}$ is true and $I(\mathrm{e})=0$ otherwise.

A learning problem is represented by a pair
$(\calH, \ell)$ of a hypothesis class
$\calH \subseteq \{h: \calX \rightarrow \calY\}$
and a loss function 
$\ell: \calX \times \calY \times \calH \rightarrow \Real$
for some input space $\calX$ and output space $\calY$.
A learner receives a sample $S=((x_1, y_1), \ldots, (x_n, y_n))$ where
each input-output pair $(x_i, y_i)$ is drawn i.i.d. according to an unknown distribution $\calD$ over $\calX \times \calY$.
The goal of the learner is to find, with high probability, a hypothesis $h \in \calH$ so that the generalization
risk $R_\calD(h) = \mathbb{E}_{(x,y)\sim \calD} \ell(x, y, h)$ is small. 
For a learning problem $(\calH, \ell)$, we define a class of loss functions as
$\calhatH = \{(x,y) \mapsto \ell(x,y,h) \mid h \in \calH \}$
when the underlying loss function $\ell$ is clear from the context.
We give the definition of the empirical Rademacher complexity,
which is used to bound the generalization risk.
\begin{defi}
[Empirical Rademacher complexity~\citep{Bartlett:2003:RGC}]
Given a sample $S=((x_1,y_1), \dots, (x_n,y_n)) \in (\calX \times \calY)^n$, 
the empirical Rademacher complexity $\Rdm_S(\calhatH)$ of a class
$\calhatH$ w.r.t.~$S$ is defined as
 $\Rdm_S(\calhatH)=\frac{1}{n}\mathbb{E}_{\vecsigma}\left[
\sup_{g \in \calhatH}\sum_{i=1}^n \sigma_i g(x_i,y_i)
 \right]$,
 where $\vecsigma \in \{-1,1\}^n$ and each $\sigma_i$ is an independent uniform random variable taking values in $\{-1,+1\}$.
\end{defi}
\paragraph{Generalization risk bound~\citep{mohri2018foundations}}
Let $(\calH,\ell) $ be a learning problem and $S$ be a sample of size $n$
drawn according to a distribution $\calD$.
Then, it holds with probability at least $1-\delta$ that
for all $h \in \calH$,
\begin{align}
    \label{align:genbound}
    R_{\calD}(h) \leq \emR_{S}(h) + 2\Rdm_S(\calhatH) + 3\sqrt{\nicefrac{\log (\nicefrac{2}{\delta})}{2n}},
\end{align}
where $\emR_{S}(h) = \frac{1}{n}\sum_{i=1}^n \ell(x_i,y_i,h)$ denotes the empirical risk of $h$ for sample $S$.

\section{Reduction scheme for ERM}
We propose a general reduction scheme for empirical risk minimization and provide useful theoretical results.
\par
\begin{defi}[{ERM-reduction}]
\label{def:erm-reducible}
  A learning problem $(\calH, \ell)$ over input-output space $\calX \times \calY$ is \emph{ERM-reducible} to another learning problem $(\calH', \ell')$
  over input-output space $\calX' \times \calY'$
  if there exist polynomial-time computable functions
  $\alpha: \calX \times \calY \rightarrow \calX' \times \calY'$ and
  $\beta: \calH' \rightarrow \calH$
  such that for any $(x, y) \in \calX \times \calY$ and for any $h' \in \calH'$,
  \begin{align}
    \label{align:reduce_condition}
    \ell(x, y, h) = \ell'(x', y', h'),
  \end{align}
  where $(x', y') = \alpha(x, y)$ and $h=\beta(h')$.
\end{defi}

Here we show the remarkable relationship between the original problem and the reduced problem.
\begin{prop}
  \label{prop:main}
  Suppose that $(\calH, \ell)$ is ERM-reducible to $(\calH', \ell')$ with transformations $\alpha$ and $\beta$.
  For any sample $S=((x_1, y_1), \ldots, (x_n, y_n)) \in (\calX \times \calY)^n$,
  the following holds:
  \begin{enumerate}[label=(\roman*)]
  \item (In)equality of the ERMs: 
  \begin{align}
    \label{align:erm_reduction_standard}
    \min_{h \in \calH} \emR_S(h) \leq &\min_{h \in \calH_\beta} \emR_S(h) \\ 
    = & \min_{h' \in \calH'}\emR_{S'}(h'),
  \end{align}
  where $\calH_\beta = \{\beta(h') \mid h' \in \calH'\}$ and
  $S' = ((x'_1,y'_1),\ldots,(x'_n,y'_n))$ with
  $(x_i', y_i') = \alpha(x_i, y_i)$ for $i \in [n]$.
\item Empirical Rademacher complexity preserving:
  \begin{align}
    \Rdm_S(\calhatH_\beta) = \Rdm_{S'}(\calhatH'). 
  \end{align}
  \end{enumerate}
\end{prop}
We can design a reduction scheme in a straightforward way as follows.
When given a sample $S$ of the original problem, 
we construct $S'$ of the reduced problem by $\alpha$ and obtain $h'$ by solving the ERM of the reduced problem. Then, we obtain the final hypothesis $h$ by $\beta$.
\par

We derive the following generalization risk bound using the propositions
on the empirical Rademacher complexity.
\begin{coro}
\label{coro:risk_bound_reduced}
Let $S = ((x_1,y_1),\ldots, (x_n, y_n))$ be a sample 
i.i.d. drawn according to unknown distribution $\calD$ in an original problem $(\calH, \ell)$.
If $(\calH, \ell)$ is ERM-reducible to $(\calH', \ell')$,
for $S'= (\alpha(x_1, y_1), \ldots, \alpha(x_n, y_n))$ and $h = \beta(h')$,
the following generalization risk bound holds
with a probability at least $1-\delta$ for all $h \in \calH_\beta$:
\begin{align}
    R_\calD(h) \leq \emR_{S'}(h') 
     + 2\Rdm_{S'}(\calhatH')
     + 3\sqrt{\nicefrac{\log (\nicefrac{2}{\delta})}{2n}}.
\end{align}
\end{coro}
That is, we can guarantee the generalization bound of the original problem
because of the preservation of the empirical Rademacher complexity.

\section{MIL-Reduction framework}
This section is the highlight of this paper.
We define the ERM-reducibility to \GMIL and show the reducible condition.
Moreover, we show that some theoretical analyses can be simplified.
We use some symbols with prime (e.g., $\calX'$)
to indicate that the MIL is the reduced problem.
\subsection{Problem formulation of MIL}
Let $\calZ \subseteq \Real^{d'}$ be the instance space. $\calX' \subseteq 2^\calZ$ is an input space and
a \emph{bag} $x' \in \calX'$ is a finite set of instances chosen from $\calZ$.
Let $\calY' = \{-1, 1\}$ be an output space.
Following the formulation by~\citep{Sabato:2012:MLA}, we define, for the rest of the paper, a MIL problem as a pair $(\calH', \ell')$ of a
hypothesis class $\calH'$ and a loss function $\ell'$ of the form:
\begin{align}
\label{align:mil_H}
\calH'\!&=\!\{h': x'\!\mapsto \Psi_p(\{f_2(g(z))\!\mid z\!\in x'\})\! \mid\! g \in \calG\}, \\
\label{align:mil_l}
\ell' &: (x',y',h') \mapsto f_1(y'h'(x')),
\end{align}
where $\calG \subseteq \{g: \calZ \to \Real\}$,
$f_1: \Real \to [0,1]$ is an $a$-Lipschitz function, 
$f_2: \Real \rightarrow [-1, 1]$ is a $b$-Lipschitz function, and 
$\Psi_p: 2^{[-1,1]} \rightarrow [-1,1]$ is a $p$-norm like function, which is defined for any $p \in [1,\infty)$ as
\begin{align}
\Psi_p(V) = \left(\frac{1}{m} \sum_{i=1}^m \left(v_i + 1 \right)^p \right)^{1/p}\! -1
\end{align}
for every finite set $V=\{v_1,v_2,\dots,v_m\} \subseteq 2^{[-1,1]}$.
We define $\Psi_\infty$ as $\lim_{p\rightarrow \infty} \Psi_p$. 
Note that $\Psi_p$ is $1$-Lipschitz for any $p$ \citep[see,][]{Sabato:2012:MLA}. 
In MIL tasks, $\Psi_p$ is a user-defined function and behaves as an aggregation of some bag information. Typical $\Psi_p$ are the $\max$ operator ($p=\infty$) and average ($p=1$).

The only difference in the hypothesis of~\citep{Sabato:2012:MLA}
is $f_2$. $f_2$ appears redundant (because $f_2 \circ g$ can be replaced by a single function) but plays an important role in the reduction (the examples are shown in Section~\ref{sec:examples}).

Here we give the definition of ERM-reducibility in a straightforward way.
\begin{defi}[MIL-reducibility]
A learning problem $(\calH, \ell)$ is said to be \GMIL-reducible if there exists a
MIL problem $(\calH', \ell')$ such that $(\calH, \ell)$ is ERM-reducible to $(\calH', \ell')$.
\end{defi}
Hereinafter, the scheme for ERM-reduction to MIL is called \emph{MIL-reduction scheme}.  

\subsection{Rademacher complexity bound}
We show the empirical Rademacher complexity bound for 
the \GMIL-reducible problems using our reduction scheme.
As aforementioned, the main advantage of our reduction scheme is to 
allow us to apply the empirical Rademacher complexity bound of 
the reduced problem to the original problems.
In this paper, we utilize the bound provided by~\citet{Sabato:2012:MLA}.
\begin{theo}[An application of Theorem 20 of~\citep{Sabato:2012:MLA}]
\label{theo:sabato}
Let $(\calH', \ell')$ be a MIL problem defined in Eq.\eqref{align:mil_H} and
\eqref{align:mil_l}.
Let $S' = ((x'_1, y'_1), \ldots, (x'_n, y'_n))$ be a sample
with average bag size $r_{S'}$.
Let $\calhatG = \{f_2 \circ g \mid g \in \calG\}$.
If there exist $C, \rho \geq 0$ such that for all sufficiently large $n$,
\begin{align}
\Rdm_{S'}(\calhatG) \leq \frac{C\ln^\rho(n)}{\sqrt{n}},
\end{align}
then 
\begin{align}
    \Rdm_{S'}(\calhatH') = O \left(
    \frac{
    \log \left(a^2 n^2 r_{S'} \right) 
    \left(\frac{aC}{\rho+1}\ln^{\rho+1}(a^2n) \right)
    }
    {\sqrt{n}}
    \right),
\end{align}
where $\calhatH' = \{\hat{h}': x' \mapsto f_1(y' h'(x')) \mid h' \in \calH' \}$.
\end{theo}
As mentioned in~\citep{Sabato:2012:MLA}, 
we obtain the following bound when $\calG$ is a set of linear functions.
\begin{coro}
Let $\calG = \{g: z \mapsto \langle {w'}, z \rangle \mid w' \in \Real^{d'}, \|w'\| \leq C_1\}$ and assume that $\|z\| \leq C_2$.
Then, the following bound holds:
\begin{align}
\label{coro:risk_bound_linear}
    \Rdm_{S'}(\calhatH) = O \left(
    \frac{
    \log \left(a^2  n^2 r_{S'} \right) 
    \left({abC_1C_2}\ln(a^2n) \right)
    }
    {\sqrt{n}}
    \right).
\end{align}
\end{coro}
The above bound is easily derived from 
the result of $\Rdm_{S'}$ \citep[see the proof of Theorem 20 of][]{Sabato:2012:MLA}) and
$\Rdm_{S'}(\calhatG) \leq b\Rdm_{S'}(\calG) \leq \nicefrac{bC_1C_2}{\sqrt{n}} = \nicefrac{bC_1C_2 \ln^0(n)}{\sqrt{n}}$ \citep[see, e.g., Theorem 5.8 and 5.10 of ][]{mohri2018foundations}.
\par
Using Theorem~\ref{theo:sabato} and Corollary~\ref{coro:risk_bound_reduced}, we obtain a generalization risk bound for \GMIL-reducible problems.

\subsection{Learning algorithm}
\label{subsec:algo}
We show that, under mild conditions, the ERM of MIL becomes a convex  or a DC (Difference of Convex) programming problem.
Suppose that 
$\calG$ is a set of linear functions:
\begin{align}
\label{align:class_g_linear}
\calG = \{g: z \mapsto \langle w', z \rangle \mid w' \in \Real^{d'}, \|w'\| \leq C_1\}.    
\end{align}
Let $S' = ((x'_1, y'_1), \ldots, (x'_n, y'_n))$.
The ERM of \GMIL is formulated as follows:
\begin{align}
\label{align:erm_optprob}
    \min_{\|w'\| \leq C_1} \! \lambda \|w'\|^2
    \!+\! \sum_{i=1}^n f_1
    \left(
    y'_i
    \Psi_p \left( \left\{
    f_2\left(
    \langle w', z \rangle \mid z \in x'_{i}
    \right)
    \right\} \right)
    \right).
\end{align}
For the optimization problem~\eqref{align:erm_optprob}, 
we show that the following propositions hold.
\begin{prop}
\label{prop:poly}
If 
$y_i'=-1$ for any $i \in [n]$
for sample $S'$,
$f_1$ is convex and nonincreasing~\footnote{More precisely, the extended-value extension $f_1$ also must be nonincreasing (See details in~\citep{boyd-vandenberghe:book04}).}, 
and $f_2$ is a nondecreasing convex function,
and $\calG$ is given as~Eq.\eqref{align:class_g_linear},
then the ERM of
$(\calH', \ell')$ is a convex programming problem.
\end{prop}

\begin{prop}
\label{prop:DC}
If 
$f_1$ is a nonincreasing convex~\footnotemark[1] 
and $f_1(c)$ is a homogeneous function of degree $1$
for $c \in [-1, 1]$\footnote{For example, hinge-loss function $f(c)= \max\{0, 1-c\}$ satisfies this condition.}, and
$f_2$ is a nondecreasing convex function,
and $\calG$ is given as~Eq.\eqref{align:class_g_linear},
then ERM of
$(\calH', \ell')$ is a DC programming problem.
\end{prop}
Generally, it is difficult to find a global minimum
for a DC programming problem; however, it is known that we can find a solution with $\epsilon$-approximation of local optima~\citep[see, e.g.,][]{le2018dc}.
We introduce a standard DC algorithm to solve \eqref{align:erm_optprob} in Algorithm~\ref{alg:DCA} in Sec.~\ref{sec:appendix_DCA}.
\par
The propositions indicate that, if $(\calH, \ell)$ is \GMIL-reducible to 
$(\calH', \ell')$ and satisfies either of the above conditions, 
then the solution $h \in \calH_\beta$
in the original problem can be obtained by a unified learning algorithm.

\section{\GMIL-reducible Examples}
\label{sec:examples}
In this section, we demonstrate that various learning problems
can be reduced to MIL by the proposed reduction scheme.
The results imply that our MIL-reduction gives a unified scheme
for designing and analyzing learning algorithms for various learning problems~\footnote{The reduction of multi-task learning and top-1 ranking learning negative feedback are shown in~Sec.\ref{sec:multi-task} and \ref{sec:trl_neg}
owing to space limitations.}.
\subsection{The existing problems}
\subsubsection{Multi-class learning problem}
\label{subsec:mcl}
{\bf Problem setting:}
Let $\calX \subseteq \Real^d$ be an instance space, and $\calY=[k]$ be an output space.
The learner receives the set of labeled instances
$S = ((x_1,y_1), \ldots, (x_n, y_n)) \in (\calX \times \calY)^n$,
where each instance is drawn i.i.d. according to some unknown distribution $\calD$.
The learner predicts the label of $x$ using the hypothesis
$h \in \calH = \{x \mapsto \arg\max_{j \in [k]} \langle w_j, x \rangle \mid \forall j\in[k],w_j \in \Real^d \}$. 
Let $\ell: (x, y, h) \mapsto \Gamma(\langle w_y, x\rangle - \max_{j \in \calY \backslash y} \langle w_j, x \rangle)$ be a loss function, 
where $\Gamma: \Real \rightarrow [0,1]$ is a convex, 
nonincreasing and $a$-Lipschitz function.
The generalization risk and empirical risk of $h$ are defined as:
\begin{align}
  R_\calD(h)\!=\!\Expo_{(x,y)\sim \calD}\! \ell \left(x,y, h \right),
  \emR_S(h)\!=\! \frac{1}{n}\sum_{i=1}^n\! \ell \left(x_i,y_i,\! h \right).
\end{align}
We obtain the following by using MIL-reduction scheme:
\begin{theo}
\label{theo:mcl_reduction}
Multi-class learning problem is \GMIL-reducible.
\end{theo}
\begin{proof}
For any $(x,y)$, we define
\begin{align}
  \label{align:dk_z}
\eta_{(x,y)} = (\bzero, \ldots, \bzero, \underbrace{x}_{y\mathrm{-th~block}}, \bzero, \ldots, \bzero),
\end{align}
where $\bzero$ is a $d$-dimensional vector, the elements of which are all $0$.
On the \GMIL-reduction scheme, 
suppose that
$p=\infty$; $f_1(c)=\Gamma(2cC_1C_2)$, $f_2(c)=c/2C_1C_2$ (shifting function to $[-1,+1]$); $\alpha(x,y) = (x'_{(x,y)}, y')$ where $x'_{(x,y)}=\{\eta_{(x,j)} - \eta_{(x,y)} \mid \forall j \in \calY \backslash y\}$; $y'=-1$; for any $z \in \Real^{kd}$,
$\calG=\{g: z \mapsto \langle (w'_1, \ldots, w'_k), z \rangle \mid 
w'_j \in \Real^d, \forall j\in [k], \|W'\| \leq C_1 \}$ where 
$W'=(w'_1, \ldots, w'_k)$ and
$\|W'\| = \sqrt{\sum_{j=1}^k\|w'_j\|^2}$; $\beta(h'): x \mapsto \arg\max_{j \in [k]} \langle w'_j, x \rangle$.
Then, for any $(x,y)$ and $h \in \calH$,
\begin{align}
&\ell'(x',y',h')
=
 f_1\left(
 y' \Psi_p\left( 
 \left\{
 f_2 \left(
 g(z) \mid z \in x'_{(x,y)}
 \right\}
 \right)
 \right)
 \right)
 \\
 &=
\Gamma \left(-\frac{1}{2C_1C_2} \Psi_\infty \left(\left\lbrace 2C_1C_2 \left( g(z) \mid z \in x_{(x,y)}'\right\rbrace \right)\right)\right)\\
&=  
\Gamma \left(-\frac{1}{2C_1C_2} \max \left(2C_1C_2 \left\lbrace \left( g(z) \mid z \in x_{(x,y)}' \right\rbrace \right) \right) \right)\\
&=
\Gamma\left(-\frac{2C_1C_2}{2C_1C_2} \max \left(\left\lbrace \left( g(z) \mid z \in x_{(x,y)}'\right\rbrace\right)\right)\right)\\
&= 
\Gamma\left(
 -\left(\max
 \left\{ \langle w', \eta_{(x,j)} - \eta_{(x,y)}\rangle \mid \forall j \in \calY \backslash y \right\} \right) 
 \right)\\
 &= \Gamma\left(
 -\left(\max_{j \in \calY\backslash y}
 \left(\langle w_j, x \rangle - \langle w_y, x \rangle\right) \right)
 \right)\\
 &= 
 \ell(x, y, h) 
 \end{align}
\end{proof}
\par
The empirical Rademacher complexity is immediately derived as follows by
observing the reduction process.
\begin{coro}
\label{coro:mc_bound}
We assume that $\|x_i\| \leq C_2$ for any $i \in [n]$.
In the reduced MIL problem from multi-class learning problem,
the empirical Rademacher complexity of $\calhatH'$ is given as:
\begin{align}
\Rdm_{S'}(\calhatH') = O \left(
    \frac{
    \log \left(\hata^2 2n^2 (k-1) \right) 
    \left({2\hata}\ln(\hata^2n) \right)
    }
    {\sqrt{n}}
    \right),
\end{align}
where $\hata=2aC_1C_2$ and we assume $\|w'\| \leq C_1$ in the reduced MIL.
\end{coro}
We used the fact that the bag size is $(k-1)$ for all $x_i'$ (i.e., $r_{S'}=k-1$)
and $\Rdm(\calhatG) \leq \nicefrac{2}{\sqrt{n}}$ by setting $f_2(c)=\nicefrac{c}{C_1C_2}$. 
Using Corollary~\ref{coro:risk_bound_reduced}, we can obtain the generalization
risk bound for the multi-class learning.
\par
The learning algorithm is obtained by the following result.
\begin{coro}
The reduced ERM of the MIL from multi-class learning is a convex programming problem.
\end{coro}
The proof of Theorem~\ref{theo:mcl_reduction} shows that
$f_2$ is nondecreasing convex and $y_i'=-1$ for all $i \in [n]$.
Therefore, by Proposition~\ref{prop:poly}, if we consider $\Gamma$ that is a nonicreasing and convex function, the ERM of the reduced MIL problem is 
a convex programming problem and solved in polynomial time.

\subsubsection{Complementarily labeled learning problem}
\label{subsubsec:complementarily}
  Complementarily labeled learning was proposed by~\citet{ishida2017learning}.
 In this problem, some training instances are complementarily labeled (e.g., instance $x_i$ is NOT $y_i$).
 We essentially follow the problem setting and some assumptions provided by~\citet{ishida2017learning}. 
\par
{\bf Problem setting:}
Let $\calX \subseteq \Real^d$ be an instance space and $\calY=[k]$ be an output space.
Let $\calD$ be an unknown distribution over $\calX \times \calY$.
We assume that the learner receives a sample $S$ drawn i.i.d. according to the distribution $\calD'$
which provides the true label with unknown probability $\theta$ and the complementary label with unknown probability $1-\theta$.
Moreover, we assume that the complementary label is chosen with a uniform probability
(i.e., all complementary labels are equally chosen with the probability $1/(k-1)$).~\footnote{This assumption was proposed by~\citet{ishida2017learning} as a reasonable scenario in some practical tasks (e.g., crowdsourcing).}
More formally, we assume that the sample is given as
$S = ((x_1,y_1, \gamma_1) \ldots, (x_n, y_n, \gamma_n))$ 
which is drawn i.i.d. according to the distribution $\calD'$ over $\calD \times \{\F, \T\}$,
where $\gamma_i=\T$ means that $y_i$ is the true label
and $\gamma_i=\F$ means that $y_i$ is the complementary label (i.e., it indicates that $x_i$ is NOT $y_i$).
For any $(x, y) \sim \calD$, $\calD'(x, y, \T) = \theta$ and $\calD'(x, \bar{y}, \F) = \frac{1-\theta}{k-1}$ for any $\bar{y} \neq y$ (i.e., the complementary label is chosen with a uniform probability).
The other basic settings are the same as those for the aforementioned multi-class learning.
The learner predicts the label of $x$ using the hypothesis
$h \in \calH = \{x \mapsto \arg\max_{j \in [k]} \langle w_j, x \rangle \mid \forall j\in[k], w_j \in \Real^d \}$. 
The final goal of the learner is to find $h \in \calH$ with a small multi-class classification risk:
\begin{align}
  \RMC_\calD(h) = \Expo_{(x,y)\sim \calD} I \left(y \neq h(x) \right).
\end{align}
However, it is difficult to minimize the empirical multi-class classification risk
directly using the complementarily labeled data.
Therefore, we consider the following risk\footnote{\citet{ishida2017learning} used a different surrogate risk. However, they and we have a common goal: to minimize $\RMC_{\calD}(h)$.}.
\begin{align}
  \label{align:|cl_origin}
  \RLC_{\calD'}(h)= 
  \Expo_{(x, y, \gamma)\sim \calD'}
  \left[I\left( \gamma = (y \neq h(x))  \right)  \right].
\end{align}
This risk implies that when $\gamma = \T$,
the learner does not incur a risk if it predicts the true label.
When $\gamma = \F$, the learner does not incur a risk if it
predicts an assigned nontrue label.
Thus, the risk measure is defined using the pair $(y, \gamma) \in (\calY \times \{\F, \T\})$.
We can show that achieving a small $\RLC_{\calD'}(h)$ is consistent with 
achieving small $\RMC_{\calD}(h)$ as follows:
\begin{lemm}
  \label{lemm:comp_gen}
For any $h \in \calH$, $\RMC_{\calD}(h) = \frac{k-1}{\theta(k -2)+1}\RLC_{\calD'}(h)$ holds.
\end{lemm}
Thus, minimizing $\RLC_{\calD'}(h)$ is a reasonable way to
achieve a high multi-class classification accuracy.
\par
Generally, there is no loss function 
$\ell((x,\gamma),y,h)$ which is a convex upper bound on 
the zero-one loss $I\left( \gamma = (y \neq h(x))  \right)$
over the domain $w$. This is because if $I(\gamma=\T)=1$ then $\max$
is convex w.r.t. $w$; however, if $I(\gamma=\T)=-1$ then $-\max=\min$ is concave w.r.t. $w$.
Therefore, we consider the convex upper bounded loss only on the risk for complementarily labeled data (i.e., the concave risk for the normally labeled data)
using $\Gamma: \Real \rightarrow [0,1]$ as $\Gamma\left(\max_{j \in \calY\backslash y}\langle (w_{j} - w_y), x\rangle\right)$.
We then define the nonconvex risk
$\ell(x,(\gamma,y),h) = \Gamma\left(I(\gamma=\T) \times \left(\max_{j \in \calY\backslash y}\langle (w_{j} - w_y), x\rangle\right) \right).$
The empirical risk is formulated as:
\begin{align}
  \label{align:er_lcl_origin}
  \emRLC_{S}(h) = \frac{1}{n}\sum_{i=1}^n
  \ell \left(x_i,(\gamma_i, y_i), h \right).  
\end{align}
\par
The following is obtained by MIL-reduction scheme.
\begin{theo}
\label{theo:cll_reduction}
Complementarily labeled learning is \GMIL-reducible.
\end{theo}
The difference from the reduction in multi-class learning is that only $y'$ takes $\{-1, 1\}$.
$y'$ behaves as a \emph{switch} that turns the loss of complementarily or normally labeled data.
\par
The empirical Rademacher complexity is bounded as:
\begin{coro}
We assume that $\|x_i\| \leq C_2$ for any $i \in [n]$.
In the reduced MIL problem from complementarily labeled learning,
the empirical Rademacher complexity of $\calhatH'$ is given by:
\begin{align}
\Rdm_{S'}(\calhatH') = O \left(
    \frac{
    \log \left(\hata^2 n^2 (k-1) \right) 
    \left({2\hata}\ln(\hata^2n) \right)
    }
    {\sqrt{n}}
    \right),
\end{align}
where $\hata=2aC_1C_2$ and we assume $\|w'\| \leq C_1$ in the reduced MIL problem.
\end{coro}
We use the same argument as in Corollary~\ref{coro:mc_bound}.
Using Corollary~\ref{coro:risk_bound_reduced} and Lemma~\ref{lemm:comp_gen}, we obtain the generalization bound for the complementarily labeled learning.
\par
The learning algorithm is derived by the following result:
\begin{coro}
The reduced ERM of the MIL from complementarily labeled learning
is a DC programming problem. If the sample contains only complementarily labeled data, the learning problem is a convex programming problem.
\end{coro}
Generally, $y' \in \{-1, 1\}$ in complementarily labeled learning.
Using the proof of Theorem~\ref{theo:cll_reduction} and 
by Proposition~\ref{prop:DC}, if we consider $\Gamma(c)$ which is a nondecreasing and homogeneous function of degree 1 for $c \in [-1,1]$ such as hinge-loss function, 
we can solve the problem by DC algorithm as shown in Algorithm~\ref{alg:DCA}.
Note that, if the sample contains only complementarily labeled data (i.e., $\forall i\in [n]$, $y_i=-1$), it becomes a convex programming problem.

\subsubsection{Multi-label learning problem}
\label{sec:multi-label}
\paragraph{Problem setting}
Let $\calX \subseteq \Real^d$ be an instance space and $\calY \in \{-1, 1\}^k$ be an output space,
and $\calD$ be an unknown distribution over $\calX$.
Unlike the standard multi-class learning setting introduced in Section~\ref{subsec:mcl}, each instance may have multiple labels 
(e.g., in text-categorization tasks, some texts have multiple topics such as IT and business).
$y^j$ denotes the $j$-th element of $y_i$.
The learner receives a labeled sample $S=(x_1, y_1), \ldots, (x_n,y_n) \in \calX \times \calY$ which is drawn i.i.d. according to the distribution $\calD$.
The learner predicts whether $x$ belongs to class $j \in [k]$ or not 
using the hypothesis
$h \in \calH = \{(x,j) \mapsto \sign (\langle w_j, x \rangle) \mid \forall w_j \in \Real^d \}$. 
Let $\ell: (x, y, h) \mapsto \frac{1}{k} \sum_{j=1}^k\Gamma(-y^j \langle w_j, x \rangle)$ where $\Gamma: \Real \rightarrow [0,1]$ is a convex, nondecreasing and $b$-Lipschitz function~\footnote{Note that we use the negative score $-y^j \langle w_j, x \rangle$ to employ a nondecreasing $\Gamma$.}. 
The generalization and empirical risk of $h$ are defined as:
\begin{align}
  R_\calD(h)\! =\!\Expo_{(x,y)\sim \calD} \left[\ell(x,\! y,\!h)\right],   \emR_S(h)\!=\!\frac{1}{n}\sum_{i=1}^n  \ell(x_i, y_i,\!h).
\end{align}
\normalsize
\paragraph{Reduction to \GMIL}
\begin{theo}
\label{theo:mll_reduction}
Multi-label learning is \GMIL-reducible.
\end{theo}
\begin{proof}
On the \GMIL-reduction scheme, suppose that $p=1$; $f_1: f_1(a) = -a$ for $a \in \Real$;
$f_2$ is $\Gamma$; $\alpha(x,y)=(x'_{(x,y)}, y')$ where $x'_{(x,y)}=\{(-y^1 x,1), \ldots, (-y^k x,k)\}$; $y'=-1$; $\calG=\{g: (z,j) \mapsto \langle w'_j, z \rangle \mid 
w'_j \in \Real^d, \forall j\in [k], \|W'\| \leq C_1 \}$ where $W' = (w'_1, \ldots, w'_k)$;  $\beta(h'): (x,j) \mapsto \sign(\langle w'_j, x \rangle)$.
For any $(x,y)$ and $h\in \calH$, we have that
\begin{align}
\ell'(x', y', h')
=&
f_1\left(
y' \Psi_p\left( 
\left\{
f_2 \left(
g(z)\right) \mid z \in x'_{(x,y)}
\right\}
\right)
\right)
\\
=&
 \frac{1}{|x'_{(x,y)}|}\sum_{(y^j x,j) \in x'_{(x,y)}} 
\Gamma \left(-\langle w_j, y^j x \rangle 
\right)\\
=&
\ell(x, y, h) 
\end{align}
\end{proof}
\par
The empirical Rademacher complexity is bounded as:
\begin{coro}
We assume that $\|x_i\| \leq C_2$ for any $i \in [n]$.
In the reduced MIL problem,
the empirical Rademacher complexity of $\calhatH'$ is given as follows:
\begin{align}
\Rdm_{S'}(\calhatH') =O\left(
    \frac{
    \log \left(2n^2 k \right) 
    \left({bC_1C_2}\ln(n) \right)
    }
    {\sqrt{n}}
    \right),
\end{align}
where $\|w'\| \leq C_1$ in the reduced MIL.
\end{coro}
We used the fact that the size of each bag is $k$. 
Using Corollary~\ref{coro:risk_bound_reduced}, we obtain the generalization
risk bound for the multi-label learning.
\par
The learning algorithm is obtained by the following result.
\begin{coro}
The reduced ERM of the MIL from multi-label learning is a convex programming problem.
\end{coro}
The proof of Theorem~\ref{theo:mll_reduction} shows that,
$f_1$ is nonincreasing and convex, and $y_i'=-1$ for all $i \in [n]$.
Therefore, by Proposition~\ref{prop:poly}, if we consider $\Gamma$ that is nondecreasing and convex, the reduced problem is 
a convex programming problem and it is solved in polynomial time.


\subsection{Application to the new problems}


\subsubsection{Multi-label learning with perfectionistic loss}
{\bf Problem setting:}
In a standard multi-label learning (see Sec.\ref{sec:multi-label}),
we consider the average prediction error (loss) with the classes.
On the other hand, we consider a \emph{perfectionistic} error in multi-label learning problem. 
More formally, we consider the following loss in a multi-label 
learning:
\begin{align}
    \ell: (x, y, h) \mapsto \max_{j \in [k]}\Gamma(-y^j \langle w_j, x \rangle),
\end{align}
 where $\Gamma: \Real \rightarrow [0,1]$ is a convex, nondecreasing and $b$-Lipschitz function.
This loss means that the learner incurs the risk
unless the learner perfectly predict the correct labels.
The generalization and empirical risks of $h$ are given as $R_\calD(h) = \mathbb{E}_{(x,y)\sim \calD} \left[\ell(x, y, h)\right]$, $\emR_S(h) =  \frac{1}{n}\sum_{i=1}^n  \ell(x_i, y_i, h)$, respectively.
\par
Using MIL-reduction scheme, we obtain the following:
\begin{theo}
\label{theo:mllp_reduction}
Multi-label learning with perfectionistic loss is \GMIL-reducible.
\end{theo}
This can be derived by the same argument with multi-label learning except for $p=\infty$ (see Sec.\ref{sec:proof_mllp_reduction}).
\par
The empirical Rademacher complexity is bounded as:
\begin{coro}
We assume that $\|x_i\| \leq C_2$ for any $i \in [n]$.
In the reduced MIL problem,
the empirical Rademacher complexity of $\calhatH'$ is given as follows:
\begin{align}
\Rdm_{S'}(\calhatH') = O\left(
    \frac{
    \log \left(2n^2 k \right) 
    \left({bC_1C_2}\ln(n) \right)
    }{\sqrt{n}}
    \right),
\end{align}
where we assume $\|w'\| \leq C_1$.
\end{coro}
Interestingly, we can have the same generalization risk bound with the standard
multi-label learning. 
\par
The learning algorithm is derived by the following result.
\begin{coro}
The reduced ERM of the MIL from multi-label learning with perfectionistic loss 
is a convex programming problem.
\end{coro}
This is easily obtained by observing the reduction process shown in Sec.\ref{sec:proof_mllp_reduction} and using Prpoposition~\ref{prop:poly}.
\par
A naive approach for the multi-label learning with perfectionistic loss
is to reduce to multi-class learning.
That is, we consider all combinations of the multi-label as multi-classes 
and solve $2^k$-class learning problem with high computational cost.
However, by the above corollary, multi-label learning with perfectionistic loss
can be solved efficiently.

\subsubsection{Top-1 ranking learning}
\label{subsec:trl}
Learning to rank is a fundamental problem, and many applications, such as recommendation systems, exist.
We consider the following natural scenario in a recommendation problem;
a learner has a set that contains several items, and it wishes to
recommend an item to a target user from the set.
\par
{\bf Problem setting:} 
Let $\calV \subseteq \Real^d$ be an instance space,
and $s: \calV \rightarrow \Real$ be a target scoring function.
Let $\calX \subseteq s^\calV$ be an input space and
set $x \in \calX$ be a finite set of instances selected from $\calV$. 
The learner receives the sequence of the sets of items and the chosen item
$S=(x_1, v^*_1), \ldots, (x_n, v^*_n)$, where each $v^*_i \in x_i$ is the highest-valued item
determined by the target function $s$. 
$k$ denotes the average size of the item sets in $S$, that is, $k=\frac{1}{n}\sum_{i=1}^n |x_i|$. 
Each sample set of items is drawn i.i.d. according to an unknown
distribution $\calD$ over $2^{\calV}$.
Assume that the learner predicts the item from the item set using the hypothesis
$h \in \calH= \{x \mapsto \arg\max_{v \in x}\langle w, v \rangle \mid w \in \Real^d\}$.\footnote{We consider an $\arg\max$ with a fixed tie-breaking rule.}
Let $\ell (x, v^*, h)$ is a convex upper bound on the zero-one loss function
$I(y \neq \hat{y})$. Equivalently, we consider the zero-one loss 
$I(\langle w, v^* \rangle - \max_{v \in x\backslash v^*} \langle w, v \rangle \leq 0)$ and its convex upper bounded loss $\ell: (x,v^*, h) \mapsto \Gamma(\langle w, v^* \rangle - \max_{v \in x\backslash v^*} \langle w, v \rangle)$ where $\Gamma: \Real \rightarrow [0,1]$ is a convex, nonincreasing and $a$ Lipschitz function.
The goal of the learner is to find $h \in \calH$ with a small misranking risk
w.r.t. the target $s$. Thus, the generalization and empirical risks are formulated as follows:
\begin{align}
  R_\calD(h) \!=\!\Expo_{x \sim \calD}\!
  \left[
    \ell \left(x, v^*\!, h \right)
    \right],
  \emR_{S}(h)\!=\! \frac{1}{n} \sum_{i=1}^n \ell \left(x_i, v_i^*\!, h \right),
\end{align}
where $v^* = \arg\max_{v \in x}s(v)$.
\par
We obtain the following by using MIL-reduction scheme:
\begin{theo}
\label{theo:trl_reduction}
 Top-1 ranking learning is \GMIL-reducible.
\end{theo}
The reducible condition is satisfied when we set
$\alpha(x, v^*) = (x', y')$ where $x'= \{v - v^* \mid v \in x\backslash v^*  \}$
$y_i'=-1$ for all $i \in [n]$.
The details of the reduction process is in~Sec.\ref{sec:proof_trl_reduction}.
\par
The empirical Rademacher complexity bound is as follows:
\begin{coro}
We assume that $\|v\| \leq C_2$ for any $v \in x_i,  \forall i \in [n]$.
In the reduced MIL problem,
the empirical Rademacher complexity of $\calhatH'$ is given as follows:
\begin{align}
\Rdm_{S'}(\calhatH') = O\left(
    \frac{
    \log \left(\hata^2 n^2 (k-1) \right) 
    \left({\hata}\ln(2\hata^2n) \right)
    }
    {\sqrt{n}}
    \right),
\end{align}
where $\hata=2aC_1C_2$ and we assume $\|w'\| \leq C_1$.
\end{coro}
The generalization bound can be derived by applying
$r_{S'} = k-1$ and using the fact that $\|v\| \leq 2C_2$ for any $v \in x_i', \forall i \in [n]$ in the reduced MIL.
By using Corollary~\ref{coro:risk_bound_reduced}, we can obtain the generalization
risk bound for the Top-1 ranking learning.
\par
The learning algorithm is designed by the following result:
\begin{coro}
The reduced ERM of MIL from top-1 ranking learning
is a convex programming problem.
\end{coro}
The corollary can be easily derived from the reduction process
detailed in~\ref{sec:proof_trl_reduction}.
\par
{\bf Extension:}
We consider \emph{top-1 ranking learning with negative feedback}
which is an extension of top-1 ranking learning.
We show the details in Sec.\ref{sec:trl_neg}.
Remarkably, the ERM problem of the reduced MIL is a DC programming problem.

\section{Kernelized extension}
Although we consider a linear function set as $\calG$;
in practice, a nonlinear kernel is required for various learning tasks.
A straightforward method is to employ 
a kernel-approximation technique~\citep[see, e.g., Sec.6.6 in][]{mohri2018foundations},
which constructs feature vectors $\Phi(x) \in \Real^D$ with the theoretical guarantee that $\langle \Phi(x_1), \Phi(x_2)\rangle \approx K(x_1, x_2)$ for a user-determined dimension $D$.
However, we can use only a limited number of kernels via the approximation technique.
Therefore, we show the kernelized version of the reduction.
\subsection{Settings}
We assume that an original problem is defined by $\calH, \ell, \calX, \calY$, and
$\Phi: \calX \to \Hilbert$, where $\Hilbert$ is a reproducing kernel Hilbert space 
associated to $K(x_1,x_2) = \langle \Phi(x_1), \Phi(x_2) \rangle$.
Aside from the computability, we can virtually consider the sample as 
$S = ((\Phi(x_1), y_1), \ldots, (\Phi(x_n), y_n))$.
The ERM-reducible condition is that there exist $(x', y')=\alpha(\Phi(x), y)$, 
$h=\beta(h')$ and $\ell'$ that satisfies  
$\ell(\Phi(x),y, h) = \ell'(x_i', y_i', h')$
for any $(x,y) \in \calX \times \calY$.

Let $S' = ((x'_1, y'_1), \ldots, (x'_n, y'_n))$ and let
$\calG = \{g: z \mapsto \langle w', z \rangle \mid w' \in \Hilbert'\}$.
We assume that $(\calH, \ell)$ is MIL-reducible to $\calH', \ell'$.
The ERM of the reduced MIL is formulated as:
\begin{align}
\label{align:erm_optprob_ker}
    \min_{w' \in \Hilbert'} \lambda \|w'\|_{\Hilbert'} 
    + \calL_{w'},
\end{align}
where $\calL_{w'}=
    \sum_{i}^n f_1
    \left(
    y'_i
    \Psi_p \left( \left\{
    f_2\left(
    \langle w', z \rangle \mid z \in x'_{i}
    \right)
    \right\} \right)
    \right)$.
\subsection{Computability}
We show that the representer theorem holds for the 
optimization problem~\eqref{align:erm_optprob_ker}.
\begin{theo}[Representer theorem]
\label{theo:representer}
An optimal solution of the ERM problem~(\ref{align:erm_optprob_ker})
has the form $\tilde{w}' = \sum_{z \in P_{S'}} \mu_z z$,
where $P_{S'} = \bigcup_{i=1}^n x'_i$.
\end{theo}
Thus, the ERM problem~\eqref{align:erm_optprob_ker}
is equivalently formulated as:
\begin{align}
\label{align:erm_optprob_ker2}
    \min_{\bmu \in \Real^{|P_{S'}|}} 
    \lambda 
    \sum_{z, \hat{z} \in P_{S'}}\mu_{z}\mu_{\hat{z}}\langle z,\hat{z} \rangle
    + \calL_{\bmu},
\end{align}
where $\calL_{\bmu}\!=\!
    \sum_{i=1}^n f_1
    (y_i
    \Psi_p ( 
    \{f_2 (\sum_{z \in P_{S'}}\mu_{z} \langle z, \hat{z} \rangle ) \mid \hat{z} \in x'_i \}
    )
    )$.
\par
Therefore, if $\langle z_1,z_2 \rangle$ is polynomial-time computable for any $z_1, z_2 \in x'$ using 
the original kernel function $K$ as an oracle,
the ERM of the MIL can be solved similar to linear case according to the condition in Proposition~\ref{prop:poly} and \ref{prop:DC}
(DC algorithm for the kernel version is in Sec.~\ref{sec:DC_algorithm_ker}).
For all MIL-reducible problems introduced in the paper,
$\langle z_1,z_2 \rangle$ is polynomial-time computable using 
$K$ (see details in Sec.\ref{sec:example_ker}).
Moreover, we can construct $\beta$ in polynomial time.

\section{Discussion}
\subsection{Related work}
\label{sec:related}
{\bf Other reduction techniques:}
Several machine-learning reduction schemes exist
\citep[see, e.g.,][]{beygelzimer2015learning},
and we found general reduction schemes, such as~\citep{pitt1990prediction,beygelzimer2005error}.
A major difference between the proposed scheme and existing approaches is that 
we focus on the reduction of ERM.
Various applications of machine learning reductions, such as
reduction from multi-class learning to binary classification~\citep{james1998error, ramaswamy2014consistency},
and from ranking to binary classification~\citep{balcan2008robust,ailon2010preference,agarwal2014surrogate},
exist.
To the best of our knowledge, the reduction to MIL has not yet been discussed.
\par
{\bf Multi-Class Learning:}
Recently, \citet{Lei19} achieved $\log(k)$-dependent generalization
bound.
The proposed generalization bound is competitive with the bound.
However, our derivation is highly simpler than
the analysis of~\citep{Lei19} because the reduction allows us to
apply the existing MIL bound of~\citep{Sabato:2012:MLA}.
\par
{\bf Complementarily-labeled learning:}
\citet{ishida2017learning} provided the generalization risk bound in the case in which the training sample
contains only complementarily labeled instances (i.e., $\theta=0$). 
The proposed generalization bound is incomparable to 
the bound (see details in Sec.\ref{sec:ishida-comparison}).
\citet{ishida2017learning} selected nonconvex loss functions
and optimized the empirical risks using a gradient-based algorithm in practice.
However, there is no guarantee of the optimality of the solution.
We show that the learning problem can be solved by DC algorithm
and guarantee the local optima.
Moreover, in the special case that sample contains only complementarily labeled data, the learning problem becomes convex programming and we can obtain global optima. 
To the best of our knowledge, the provided learning algorithm is a first 
polynomial-time algorithm in the special case.
\par
{\bf Multi-label learning:}
Various approaches and generalization analyses have been provided~\citep{pmlr-v32-yu14,NIPS2015_35051070,xu2016local,xu2016robust}.
However, to the best of our knowledge, this paper is the first to propose a
$\log(k)$-dependent generalization bound for the linear (or nonlinear kernel)
hypothesis class, where $k$ is the number of classes.
\par
{\bf Multi-task learning:}
A similar generalization bound was reported by~\citep{pontil2013excess}.
Their results suggest the advantage of regularizing the weights 
$w_1,\ldots,w_T$ over $T$ tasks.
However, our result is derived from an entirely different argument
from~\citep{pontil2013excess} and the derivation is highly simplified. \par
\par
{\bf Top-1 ranking learning:}
Top-1 ranking measure was originally discussed in~\citep{hidasi2018recurrent}.
However, the basic problem setting is different from ours.
They assumed that the recommender has i.i.d. positive and negative items as the sample.
Moreover, they did not propose a general form of the problem and theoretical analysis.
\par
{\bf MIL:}
MIL was originally proposed by~\citet{Dietterich:1997}, which is
known as weakly supervised learning and there have been proposed
many real applications~\citep{Gartner02multi-instancekernels,NIPS2002misvm,pmlr-v28-zhang13a,Doran:2014,CARBONNEAU2018329}.
The generalization bound and learning algorithm have been
analyzed from the theoretical perspective~\citep{Sabato:2012:MLA,doran:thesis,suehiro2020multiple}.
There have been several studies on the relationship between MIL with other learning tasks.
\citet{zhou2007relation} showed that a classical MIL can be considered as specific semi-supervised learning.
\citet{Zhang2020RobustML} utilized MIL for extracting causal instances.
However, these works do not imply any type of reduction in the sense of computation theory: if problem A is reduced to B, then we should immediately obtain an algorithm for A from any algorithm for B combined with the reduction (input-output transformations) with a certain performance guarantee.
\citet{suehiro2020multiple} found that a local-feature-based time-series classification
problem can be reduced to a MIL problem with a generalization risk bound. However, the reduced problem is too specific.
Our results first show that various learning problems can be reduced to MIL.

\subsection{Practical implications}
An important contribution of the paper in both the theoretical and practical aspects is to provide a simple and general reduction scheme among various learning problems with theoretical guarantees on generalization bounds. This means that when faced with a new learning problem A, we can search for an existing ERM problem B that is reducible from A. If succeeded, then we immediately obtain a learning algorithm for A with a generalization bound. Usually, this process is expected to be much easier than designing a learning algorithm from scratch.
\par
In particular, we demonstrate that various learning problems are reducible to a particular problem, MIL. That is, we only have to improve ERM algorithms for MIL, which work on the original learning problems as well. Moreover, we show that ERM for MIL can be formulated as DC programming problems in Section~\ref{subsec:algo}. Therefore, we can employ a state-of-the-art DC programming package, which is rapidly evolving these days~\citep{le2018dc}. For instance, complementarily labeled learning, which is only known to have a non-convex optimization formulation~\citep{ishida2017learning,ishida19a}, would enjoy the benefits from a promising DC programming approach.
\par
\paragraph{Experiments:} We demonstrate that our theoretical results are practically useful in the following experiment on complementarily labeled learning tasks~\footnote{The code is available in \url{https://github.com/suehiro93/MIL_reduction}}.
We use three artificial datasets and four benchmark datasets available in UCI machine learning repository~\footnote{\url{https://archive.ics.uci.edu/ml/}}.
The details of artificial datasets are described in Section~\ref{sec:art_data}.
For all datasets, all training instances are complementarily labeled uniformly at random.
That is, the ERM problem which is derived from our MIL-reduction scheme becomes a convex programming problem (quadratic programming problem). On the other hand, \citep{ishida2017learning} solves a nonconvex optimization problem by using Adam~\citep{kingma2014adam}. 
The size of training sample is fixed to 1000 and we used the remaining data as a test set.
Although we did not tune the optimization hyperparameters of~\citep{ishida2017learning} (the number of epochs is 200 and the learning rate is 0.01), we stopped the learning at the epoch when the test accuracy was the maximum. The loss of~\citep{ishida2017learning} was fixed to PC loss which was the best-performed loss~\citep[see][]{ishida2017learning}.
Our regularization parameter is chosen from $\{0.01, 1, 100\}$ and the regularization parameter of \citep{ishida2017learning} is chosen from $\{0.01, 1, 100\}$.
We evaluated the average accuracy over 10 trials.

Table~\ref{tab:acc} shows that our method achieved higher classification accuracy than~\citep{ishida2017learning} on many datasets. This result indicates that our MIL-reduction scenario for complementarily labeled learning, which is derived from the proposed MIL-reduction scheme, is useful in practice.
Moreover, our ERM algorithm does not require any hyperparameters for the optimization because the optimization problem is a convex programming problem (or DC programming problem when the training sample contains both labeled and complementarily labeled instances). On the other hand, the learning algorithm provided by \citet{ishida2017learning} solves a nonconvex optimization problem and usually requires several hyperparameters (e.g., learning rate and the number of epochs) of the nonconvex-optimization solver. 
\begin{table}[t]
\centering
\caption{Average test accuracy over 10 trials.}
\label{tab:acc}
\begin{tabular}{l|cccc}
\hline
Dataset  &Class & Dim.& Ours & Ishida+  \\
\hline \hline
artificial1 & 5 & 50 & \textbf{0.9999}    & 0.9998 \\
artificial2 & 10 & 50 & \textbf{0.808}    & 0.646 \\
artificial3 & 25& 50 & 0.063    & \textbf{0.065} \\
covertype & 7 & 54 & \textbf{0.562}    & 0.549 \\
satimage & 7 & 36 & \textbf{0.804}    & 0.751 \\
waveform & 3& 40& \textbf{0.833}    & 0.832 \\
yeast & 10& 8& 0.348    & \textbf{0.407} \\
\hline
\end{tabular}
\end{table}

\subsection{Conclusion and future work}
We revealed that various learning problems can be reduced to a MIL 
problem by our ERM-based reduction scheme.
The results imply that our MIL-reduction gives a simplified and unified scheme
for the analyses for various learning problems.
Moreover, we obtained novel theoretical results for some learning problems.
A practical concern is that the applicable loss functions are limited in the current scheme. For example, some loss functions without satisfying the conditions of MIL-reducibility (e.g., square loss) cannot be used.
We explore the relaxation of the ERM-reducible condition.
An interesting open problem is how the class of MIL-reducible problems is characterized.
Our results imply that MIL is one of the hardest problems in a certain class C of learning problems. In other words, we could say that MIL is a C-complete problem. We would like to investigate how the class C is characterized.

\section*{Acknowledgment}
This work was supported by JSPS KAKENHI (Grant Number JP19H04067 and JP20H05967) and JST, ACT-X (Grant Number JPMJAX200G).

\appendix
\onecolumn


\section{Proof of Theorem~\ref{theo:sabato}}
\begin{proof}
The theorem is based on Theorem 20 of~\citep{Sabato:2012:MLA}.
Using the fact that $\psi_p$ is $1$-Lipschitz for all $p$
and $\Rdm_S$ which is shown in the proof of Theorem 20 of~\citep{Sabato:2012:MLA},
we can obtain the target theorem.
\end{proof}


\section{Proof of Proposition~\ref{prop:poly}}
\begin{proof}
First we have that $\hat{f}=f_2 \circ g$ is a convex function of $w'$ because
$f_2$ is a nondecreasing convex and $\langle w', z \rangle$ is a convex function of $w'$
(see, e.g., Eq. (3.11) in~\cite{boyd-vandenberghe:book04}).
Subsequently, we show that $\Psi_p \circ \hat{f}$ is a convex function.
Without loss of generality, we can consider $\Psi_p$ as a function $\Real^m \to \Real$ where $m$ is the size of the set $x'$. $\Psi_p$ is a nondecreasing function in each argument and $\hat{f}$ is convex and thus $\Psi_p \circ \hat{h}$ is convex.
Finally, because $-\Psi_p(\{f_2(\langle w', z \rangle) \mid z \in x' \})$ is concave and
$f_1$ is nonincreasing convex, $f_1(-\Psi_p(\{f_2(\langle w', z \rangle) \mid z \in x' \})$ is convex~\cite{boyd-vandenberghe:book04}.
\end{proof}

\section{Proof of Proposition~\ref{prop:DC}}
\begin{proof}
Because $f_1(c)$ is a homogeneous function of degree $1$ for $c \in [-1,1]$, we have
$f_1(-\Psi_p(\{f_2(\langle w', z \rangle) \mid z \in x' \})) = -f_1(\Psi_p(\{f_2(\langle w', z \rangle) \mid z \in x' \}))$.
As we proved in Proof of Proposition~\ref{prop:poly}, $f_1(-\Psi_p(\{f_2(\langle w', z \rangle) \mid z \in x' \}))$ is convex. Moreover, we have $f_1(\Psi_p(\{f_2(\langle w', z \rangle) \mid z \in x' \})) = -f_1(-\Psi_p(\{f_2(\langle w', z \rangle) \mid z \in x' \}))$ 
and thus $f_1(\Psi_p(\{f_2(\langle w', z \rangle) \mid z \in x' \}))$ is concave.
Therefore, we have that $f_1(\Psi_p(\{f_2(\langle w', z \rangle) \mid z \in x' \})) + f_1(-\Psi_p(\{f_2(\langle w', z \rangle) \mid z \in x' \}))$ is a DC function.
\end{proof}

\section{DC algorithm for the reduced MIL problem}
The algorithm is shown in Algorithm~\ref{alg:DCA}. The subproblem~\eqref{align:subprob} is a convex programming problem
that can be solved in polynomial time.
\label{sec:appendix_DCA}
\begin{algorithm}[h!]
\caption{\GMIL optimization via DC Algorithm}
\label{alg:DCA}
\begin{algorithmic}[0]
\Inputs{$S'$, $\lambda$}
 \Initialize{${w'}_0 \in \mathbb{R}^{d'}$}
 \For{$t=1,\dots,$ (until convergence)}
\State Compute the subgradient:
\begin{align}
\label{align:subgrad}
    s_t \in \nabla_{w'}\left(\sum_{i:y_i=-1} f_1
    \left( \Psi_p \left(\left\{f_2 \left(\langle w', z \rangle \right) \mid z \in x'_i \right\} \right)\right)\right) 
\end{align}
\State at $w'_{t-1}$.
\State Solve the following subproblem:
\begin{align}\label{align:subprob}
w_t' \leftarrow \arg\min_{w': \|w'\| \leq C_1}
~\lambda \|w'\|^2 
+ \sum_{i:y_i=+1} f_1 
    \left( \Psi_p \left(\left\{f_2 \left(\langle w', z \rangle \right) \mid z \in x'_i \right\} \right)\right) - s_t^{\top} w'
\end{align}    
\State
\begin{align}
\end{align}
\EndFor \\
\Return{$w_t$}
\end{algorithmic}
\end{algorithm}

\section{Proof of~Lemma~\ref{lemm:comp_gen}}
\begin{proof}
  Based on the assumption of $\calD'$, the expected risk $\RLC_{\calD'}(h)$ is represented using $\calD$, $k$, and $\theta$
  as follows:
  \begin{align}
    \label{align:rlcd}
    \RLC_{\calD'}(h) = \Expo_{(x, y)\sim \calD}
     \left[\theta I\left((y \neq h(x))  \right) + 
      (1- \theta)\sum_{\bar{y}\neq y}\frac{1}{k-1}
      I\left(\bar{y}=h(x)  \right).
      \right]    
  \end{align}
  Let  $\rho_1=I \left(y \neq h(x) \right)$ in $\RMC_{\calD}(h)$ and
  let $\rho_2=\theta I\left((y \neq h(x))  \right) +  (1- \theta)\sum_{\bar{y}\neq y}\frac{1}{k-1}  I\left((\bar{y}=h(x))  \right)$ in $\RLC_{\calD'}(h)$.
  We consider two cases of $h$ for any $h \in \calH$ as follows:
  For a fixed $(x, y)$,
  (i) If $h(x) = y$: $\rho_1=0$ and $\rho_2=0$, and thus there is no gap. 
  (ii) If $h(x) \neq y$:,  the first term of $\rho_2$
  is $\theta$ and the second term is equal to $(1-\theta)/(k-1)$,
  because there exists a unique $\hat{y}:\hat{y} \neq y$ that satisfies $\hat{y} = h(x)$.
  Therefore, $\rho_2$ is equal to $\theta + \frac{1-\theta}{k-1}$.
  In this case, $\rho_1 = 1$.
  Thus, we have the bound $\frac{k-1}{\theta(k -2)+1}\RLC_{\calD'}(h) =  \RMC_{\calD}(h)$.
\end{proof}

\section{Proof of~Theorem~\ref{theo:cll_reduction}}
\begin{proof}
We use $\eta_{(x,y)}$ defined in~(\ref{align:dk_z}).
On the \GMIL-reduction scheme, suppose that 
$p=\infty$; $f_1(c)=\Gamma(2cC_1C_2)$; $f_2(c)=c/2C_1C_2$ (shifting function to $[-1,+1]$); $\alpha(x,(\gamma,y)) = (x'_{(x,y)}, y')$ 
where $x'_{(x,y)}=\{\eta_{(x,j)} - \eta_{(x,y)} \mid \forall j \in \calY \backslash y\}$; $y'=I(\gamma=\T)$; for any $z \in \Real^{kd}$,
$\calG=\{g: z \mapsto \langle (w'_1, \ldots, w'_k), z \rangle \mid 
w'_j \in \Real^d, \forall j\in [k], \|W'\| \leq C_1 \}$ where $W'=(w'_1, \ldots, w'_k)$ and $\|W'\| = \sqrt{\sum_{j=1}^k\|w'_j\|^2}$; $\beta(h'): x \mapsto \arg\max_{j \in [k]} \langle w'_j , x \rangle$.
Then, for any $(x,y)$ and $h \in \calH$,
\begin{align}
\ell'(x', y', h')
=&
f_1\left(
y' \Psi_p\left( 
\{
f_2 \left(
g(z) \mid z \in x'_{(x,y)}
\}
\right)
\right)
\right)
\\
=&
\Gamma\left(
I(\gamma=\T) \times \Psi_\infty\left( 
\{
g(z) \mid z \in x'_{(x,y)}
\}
\right)
\right)
\\
=& \Gamma\left(
I(\gamma=\T) \times \left(\max_{j \in \calY\backslash y}
\left(\langle w_j, x \rangle - \langle w_y, x \rangle\right) \right)
\right)\\
=& 
\ell(x, (\gamma, y), h).
\end{align}
\end{proof}

\section{Multi-task learning problem}
\label{sec:multi-task}
In multi-task learning, the learner finds a common rule in
multiple-tasks,
which correctly predicts the outputs of the instances.
For example, in the multi-classification-task problem, there are three different binary classification tasks for image data, cat or dog, car or train, and apple or tomato.
\par
\paragraph{Problem setting}
Let $\calX \subseteq \Real^d$ be an input space and $\calY \in \{-1,1\}$ be an output space.
We assume that the learner has $T$ different tasks with different data
distributions.
The learner receives $T$ sets of samples $S=S_1, \ldots, S_T$ where
$S_t = ((x_1^t, y_1^t), \ldots, (x_n^t, y_n^t))$ is drawn i.i.d. according to unknown distribution $\calD_t$.
$(x^t, y^t)$ denote an instance and its label, respectively.
Let $\calH=\{h: (x^t) \mapsto \sign(\langle w_t, x^t \rangle) \mid w_t \in \Real^d \rangle\}$ be a hypothesis class. 
Let $\ell: ((x^1,\ldots, x^T), (y^1, \ldots, y^T), h) \mapsto \frac{1}{T} \sum_{t=1}^T\Gamma(-y^t \langle w_t, x^t \rangle)$ where $\Gamma: \Real \rightarrow [0,1]$ is a convex, nondecreasing and $b$-Lipschitz function.
The generalization risk and empirical risk are formulated as:
\begin{align}
  &\Expo_t[R_{\calD_t}(h)] = \frac{1}{T}\sum_{t=1}^T \Expo_{(x^t,y^t)\sim \calD_t} 
  \left[\Gamma(-y^t \langle w_t, x^t \rangle)\right], \\
  &\emR_S(h) =  \frac{1}{T}\sum_{t=1}^T \frac{1}{n}\sum_{i=1}^{n} \Gamma(-y^t_i \langle w_t, x^t_i \rangle)
  =\frac{1}{n}\sum_{i=1}^{n} \ell \left((x^1_i,\ldots, x^T_i), (y^1_i, \ldots, y^T_i), h \right).
\end{align}
\paragraph{Reduction to \GMIL}
\begin{theo}
\label{theo:mtl_reduction}
Multi-task learning is \GMIL-reducible.
\end{theo}
\begin{proof}
For simplicity, we denote $(x^1,\ldots, x^T)$ by $\bx$ and denote $(y^1, \ldots, y^T)$ by $\by$.
On the \GMIL-reduction scheme, suppose that $p=1$; $f_1: f_1(a) = -a$;
$f_2$ is $\Gamma$; $\alpha(\bx, \by)=(x'_{(\bx, \by))}, y')$ where $x'_{(\bx, \by)}=\{(y^1 x^1,1), \ldots, (y^T x^T, T)\}$; $y'=-1$; $\calG=\{g: (z,t) \mapsto \langle w'_t, z \rangle \mid 
\forall j\in [T], w'_t \in \Real^d~\mathrm{and}~ \|W'\| \leq C_1 \}$ where $W' = (w'_1, \ldots, w'_T)$; $\beta(h'): (x^t) \mapsto \sign(\langle w'_t, x^t) \rangle$.
For any $((x^1,\ldots, x^T), (y^1, \ldots, y^T))$ and $h\in \calH$, we have that
\begin{align}
\ell'(x', y', h')
=&
f_1\left(
y' \Psi_p\left( 
\left\{
f_2 \left(
g(z)\right) \mid z \in x'_{(\bx, \by)}
\right\}
\right)
\right)
\\
=&
 \frac{1}{|x'_{(\bx, \by)}|}\sum_{(x,t) \in x'_{(\bx, \by)}} 
\Gamma \left(-\langle w_t, y^t x^t \rangle 
\right)\\
=&
\ell((x^1,\ldots, x^T), (y^1, \ldots, y^T), h) 
\end{align}
\end{proof}
\paragraph{ERM algorithm}
\begin{coro}
The reduced ERM of the MIL from multi-task learning is a convex programming problem.
\end{coro}
As shown in the proof of Theorem~\ref{theo:mtl_reduction},
$f_1$ is nonincreasing and $y_i'=-1$ for all $i \in [n]$.
Thus, by Proposition~\ref{prop:poly}, if we consider $\Gamma$ that is nondecreasing and convex, the reduced MIL problem is 
a convex programming problem and solved in polynomial time.
\paragraph{Generalization bound}
 \begin{coro}
We assume that $\|x^t_i\| \leq C_2$ for any $i \in [n]$ and $t \in [T]$.
In the reduced problem,
the empirical Rademacher complexity of $\calhatH'$ is given as follows:
\begin{align}
\Rdm_{S'}(\calhatH') = O\left(
    \frac{
    \log \left(2n^2 T \right) 
    \left({bC_1C_2}\ln(n) \right)
    }
    {\sqrt{n}}
    \right),
\end{align}
where we assume $\|w'\| \leq C_1$. 
\end{coro}
We can derive the above from the same argument from the proof of Theorem~\ref{theo:mtl_reduction}.
Using Corollary~\ref{coro:risk_bound_reduced}, we can obtain the generalization
risk bound for the multi-task learning problem.

\section{Proof of~Theorem~\ref{theo:mllp_reduction}}
\label{sec:proof_mllp_reduction}
\begin{proof}
On the \GMIL-reduction scheme, suppose that $p=\infty$; $f_1: f_1(a) = -a$ for $a \in \Real$;
$f_2$ is $\Gamma$; $\alpha(x,y)=(x'_{(x,y)}, y')$ where $x'_{(x,y)}=\{(-y^1 x,1), \ldots, (-y^k x,k)\}$; $y'=-1$; $\calG=\{g: (z,j) \mapsto \langle w'_j, z \rangle \mid 
w'_j \in \Real^d, \forall j\in [k], \|W'\| \leq 1 \}$ where $W' = (w'_1, \ldots, w'_k)$; $W' = (w'_1, \ldots, w'_k)$; $\beta(h'): (x,j) \mapsto \langle w'_j, x \rangle$.
For any $(x,y)$ and $h\in \calH$, we have that
\begin{align}
\ell'(x', y', h')
=&
f_1\left(
y' \Psi_p\left( 
\left\{
f_2 \left(
g(z)\right) \mid z \in x'_{(x,y)}
\right\}
\right)
\right)
\\
=&
 \max_{(y^j x,j) \in x'_{(x,y)}} 
\Gamma \left(-\langle w_j, y^j x \rangle 
\right)\\
=&
\ell(x, y, h) 
\end{align}
\end{proof}


\section{Proof of~Theorem~\ref{theo:trl_reduction}}
\label{sec:proof_trl_reduction}
\begin{proof}
On the \GMIL-reduction scheme, suppose that $p=\infty$; $f_1(c)=\Gamma(2cC_1C_2)$; $f_2(c)=c/2C_1C_2$;
$\alpha(x, v^*) = (x', y')$ where $x'= \{v - v^* \mid v \in x\backslash v^*  \}$; 
$y'=-1$; 
$\calG = \{g: z \mapsto \langle w', z \rangle \mid \|w'\| \leq C_1 \}$;
$\beta(h'): x \mapsto \arg\max_{v \in x} \langle w', v\rangle$.
For any $(x, v^*)$ and $h \in \calH$, the following holds:
\begin{align}
\ell'(x', y', h')
=&
f_1\left(
y' \Psi_p\left( 
\{
f_2 \left(
g(z) \mid z \in x'_{(x,y)}
\}
\right)
\right)
\right)
\\
=&
\Gamma\left(
- \Psi_\infty\left( 
\{
g(z) \mid z \in x'_{(x,y)}
\}
\right)
\right)
\\
=& \Gamma\left(
-\left(\max_{j \in A\backslash x^*}
\left(\langle w, v \rangle - \langle w, v^* \rangle\right) \right)
\right)\\
=& 
\ell(x, v^*, h) 
\end{align}
\end{proof}

\section{Top-1 ranking learning with negative feedback}
\label{sec:trl_neg}
As an extension of the Top-1 rank learning problem,
we consider the following scenario.
In practice, some item sets do not include the user-preferred item.
Therefore, we assume that the item sets are partitioned into two types:
the item sets that include the most preferred item
and those that do not include the preferred item.
For the second type of item set, we assume that we can receive 
information on non-preferred items as negative feedback from the user.
\par
More formally, we assume that the target user has a scoring function $s$
and a parameter $\gamma_i \in \{-1, +1\}$,
where $\gamma$ takes $+1$ for an item set that includes the preferred
item and takes $-1$ otherwise.
The learner receives the sequence of the sets of items
and the chosen item with positive or negative information
$S = (x_1, (v^*_1, \gamma_1)), \ldots, (x_n, (v^*_n, \gamma_n)$.
$\gamma_i=+1$ indicates that item set $x_i$ includes the preferred item,
and $\gamma_i=-1$ indicates that the item set $x_i$ does not include the preferred item.
For the item set $x_i$ with $\gamma=+1$, 
$v_i^* = \max_{v \in x_i} s(v)$.
Conversely, for the item set $x_i$ with $\gamma=-1$, 
$v_i^* \in \{\tilde{x}=x \backslash \tilde{v} \mid \tilde{v}=\max_{v \in x_i} s(v)\}$,
that is, if $\gamma=-1$,
the user selects an item except for the best-scored item by $s$.
Note that we assume that $\gamma$ is a known parameter 
only in the training phase.
The other settings are the same as those in Sec.~\ref{subsec:trl}.
\par
A reasonable goal of the learner is to predict the best item from a given set of items even in this setting.
Therefore, the learner can recommend the most preferred item if $\gamma=+1$ and
can recommend a preferable item if $\gamma=-1$.
Similar to top-1 ranking learning, we consider a loss function $\ell: (x,(v^*,\gamma), h) \mapsto \Gamma(\gamma (\langle w, v^* \rangle - \max_{v \in x\backslash v^*} \langle w, v \rangle))$ where $\Gamma: \Real \rightarrow [0,1]$ is a convex, nonincreasing and $a$-Lipschitz function.
The generalization risk and empirical risk are formulated as follows:
\begin{align}
  &R_\calD(h) = \Expo_{(x, \gamma) \sim \calD}
  \left[
    \ell \left(x, (v^*, \gamma), h \right)
    \right],\\
  &\emR_{S}(h)= \frac{1}{n} \sum_{i=1}^n \ell \left(x, (v_i^*, \gamma_i), h \right),
\end{align}
where $v^* = \arg\max_{v \in x}s(v)$.
\paragraph{Reduction to MIL}
\begin{theo}
\label{theo:trln_reduction}
Top-1 ranking learning with negative feedback is \GMIL-reducible.
\end{theo}
The difference from the top-1 ranking learning is just
$y_i'=-\gamma_i$, and thus we can easily prove it.
\begin{proof}
On the \GMIL-reduction scheme, suppose that $p=\infty$; $f_1(c)=\Gamma(2cC_1C_2)$; $f_2(c)=c/2C_1C_2$;
$\alpha(x, v^*) = (x', y')$ where $x'= \{v - v^* \mid v \in x\backslash v^*  \}$; 
$y'=-\gamma$; 
$\calG = \{g: z \mapsto \langle w', z \rangle \mid \|w'\| \leq 1 \}$;
$\beta(h'): x \mapsto \arg\max_{v \in x} \langle w', v\rangle$.
For any $(x, v^*)$ and $h \in \calH$, the following holds:
\begin{align}
\ell'(x', y', h')
=&
f_1\left(
y' \Psi_p\left( 
\{
f_2 \left(
g(z) \mid z \in x'_{(x,y)}
\}
\right)
\right)
\right)
\\
=&
\Gamma\left(
\gamma\left( \Psi_\infty\left( 
\{
g(z) \mid z \in x'_{(x,y)}
\}
\right)
\right)
\right)
\\
=& \Gamma\left(
\gamma\left(\max_{j \in A\backslash x^*}
\left(\langle w, v \rangle - \langle w, v^* \rangle\right) \right)
\right)\\
=& 
\ell(x, v^*, h) 
\end{align}
\end{proof}
\paragraph{Generalization bound}
\begin{coro}
We assume that $\|v\| \leq C_2$ for any $v \in x_i  \forall i \in [n]$.
In the reduced MIL problem,
the empirical Rademacher complexity of $\calhatH'$ is given as follows:
\begin{align}
\Rdm_{S'}(\calhatH') = O\left(
    \frac{
    \log \left(\hata^2 n^2 (k-1) \right) 
    \left({2\hata}\ln(\hata^2n) \right)
    }
    {\sqrt{n}}
    \right),
\end{align}
where $\hata=2aC_1C_2$ we assume $\|w'\| \leq C_1$.
\end{coro}
Using Corollary~\ref{coro:risk_bound_reduced}, we can obtain the generalization
risk bound for the Top-1 ranking learning with negative feedback.

\paragraph{ERM algorithm}
\begin{coro}
The reduced ERM of MIL from top-1 ranking learning with negative feedback
is a DC programming problem.
\end{coro}
In top-1 ranking learning, $y' \in \{-1, 1\}$.
By the proof of Theorem~\ref{theo:trln_reduction} and 
by Proposition~\ref{prop:DC}, if we consider a loss function $\Gamma(c)$ as a nondecreasing and homogeneous function of degree 1 for $c \in [-1,1]$ such as hinge-loss, 
we can solve the problem by DC algorithm as shown in Algorithm~\ref{alg:DCA}.


\section{Proof of Theorem~\ref{theo:representer}}
\begin{proof}
For the optimization problem~\eqref{align:erm_optprob_ker}, we can apply the standard representer theorem (see, e.g., Theorem 6.11 of~\cite{mohri2018foundations}).
We define $\Hilbert_1$ as the subspace 
spanned by $\{\langle z, \cdot \rangle \mid z \in P_{S'}\}$, namely, 
$\Hilbert_1=\{w \in \Hilbert \mid w = \sum_{z\in P_{S'}} \mu_z z, \mu_z \in \Real \}$.
For any $w \in \Hilbert$, we can consider the decomposition $w = w_1 + w_1^\perp$, where $w_1 \in \Hilbert_1$, and $w_1^\perp \in \Hilbert_1^\perp$ is its orthogonal component. 
Because $\Hilbert_1$ is a subspace of $\Hilbert$, 
$\|w\|_{\Hilbert}=\sqrt{\|w_1\|_{\Hilbert}^2 + \|w_1^\perp\|_{\Hilbert}^2} \geq \|w_1\|_{\Hilbert}$.
Moreover, by the definition of $\Hilbert_1$,
$\langle w, z\rangle= \langle w_1, z\rangle$.
Thus, $f_1(y_i'\Psi_p(\{f_2(\langle w,z \rangle ) \mid z \in x'_{i} \})) =f_1(y_i'\Psi_p(\{f_2(\langle w_1,z \rangle ) \mid z \in x'_{i} \}))$ and 
$\|w_1\|_\Hilbert \leq \|w\|_\Hilbert$.
This implies that the optimal solution
is contained in $\Hilbert_1$.
\end{proof}

\section{DC algorithm for kernelized extension}
The algorithm is shown in Algorithm~\ref{alg:DCA_ker}.
\label{sec:DC_algorithm_ker}
\begin{algorithm}
\caption{\GMIL optimization via DC Algorithm (kernelized)}
\label{alg:DCA_ker}
\begin{algorithmic}[0]
\Inputs{$S'$, $\lambda$}
 \Initialize{${\bmu}_0 \in \mathbb{R}^{|{P_{S'}}|}$}
 \For{$t=1,\dots,$ (until convergence)}
\State Compute the subgradient:
\begin{align}
\label{align:subgrad2}
    s_t \in \nabla_{\bmu}\left(\sum_{i:y_i=-1} f_1
    \left( \Psi_p \left(\left\{f_2 \left(\sum_{v \in P_{S'}}\mu_{v} \langle v, z\rangle \right) \mid z \in x'_i \right\} \right)\right)\right) 
\end{align}
\State at $\bmu_{t-1}$.
\State Solve the following subproblem:
\begin{align}\label{align:subprob_ker}
\nonumber
\bmu_t \leftarrow \arg\min_{{\bmu} \in \mathbb{R}^{|{P_{S'}}|}}
&~\lambda \sum_{v, \hat{v} \in P_{S'}}\mu_{v}\mu_{\hat{v}}\langle v,\hat{v} \rangle\\ 
&+ \sum_{i:y_i=+1} f_1
    \left( \Psi_p \left(\left\{f_2 \left(\sum_{v \in P_{S'}}\mu_{z} \langle z, x\rangle \right) \mid z \in x'_i \right\} \right)\right) \\
    &- s_t^{\top} \bmu
\end{align}
\EndFor \\
\Return{$\bmu_t$}
\end{algorithmic}
\end{algorithm}

\section{Example of the reduction of kernelized learning problems: multi-class learning}
\label{sec:example_ker}
\subsection{Reduction to MIL with kernel}
\begin{theo}
Multi-class learning with kernel is \GMIL-reducible.
\end{theo}
\begin{proof}
For any $(x,y)$, we define
\begin{align}
  \label{align:dk_z_ker}
\eta_{(x,y)} = (0_\Hilbert, \ldots, 0_\Hilbert, \underbrace{\Phi(x)}_{y\mathrm{-th~block}}, 0_\Hilbert, \ldots, 0_\Hilbert) \in \Hilbert^k,
\end{align}
where $0_\Hilbert$ is a point in $\Hilbert$ satisfying $\langle 0_\Hilbert, v \rangle=0$ for any $v \in \Hilbert$.
On the \GMIL-reduction scheme, 
suppose that
$p=\infty$; $f_1(c)=\Gamma(cC_1C_2)$; $f_2(c)=c/C_1C_2$; $\alpha(x,y) = (x'_{(x,y)}, y')$ where $x'_{(x,y)}=\{\eta_{(x,j)} - \eta_{(x,y)} \mid \forall j \in \calY \backslash y\}$; $y'=-1$;
$\calG=\{g: z \mapsto \langle (w'_1, \ldots, w'_k), z \rangle \mid \forall j \in [k], w'_j \in \Hilbert, \|W'\|_{\Hilbert^k} \leq C_1 \}$ where $W' = (w'_1, \ldots, w'_k)$, $\|W'\|_{\Hilbert^k} = \sqrt{\sum_{j=1}^k\|w'_j\|^2_{\Hilbert}}$.
Then, for any $(x,y)$ and $h \in \calH$,
\begin{align}
\ell'(x', y', h')
=&
f_1\left(
y' \Psi_p\left( 
\{
f_2 \left(
g(z) \mid z \in x'_{(x,y)}
\}
\right)
\right)
\right)
\\
=&
\Gamma\left(
- \Psi_\infty\left( 
\{
g(z) \mid z \in x'_{(x,y)}
\}
\right)
\right)
\\
=& \Gamma\left(
-\left(\max_{j \in \calY\backslash y}
\left(\langle W', \eta_{(x,j)} - \eta_{(x,y)} \rangle\right) \right)
\right)\\
=& \Gamma\left(
-\left(\max_{j \in \calY\backslash y}
\left(\langle w_j, \Phi(x) \rangle - \langle w_y, \Phi(x) \rangle\right) \right)
\right)\\
= &
\ell(x, y, h) 
\end{align}
\end{proof}

\subsection{Construction of $\beta$}
By Theorem~\ref{theo:representer}, $W'$ is returned by using $\mu$ as
\begin{align}
    W' = \sum_{z \in P_{S'}} \mu_{z} z.
\end{align}
Moreover, $w'_j$ can be represented as:
\begin{align}
    w'_j = \sum_{z[j] \in P_{S',j}} \mu_{z[j]} v[j],
\end{align}
where $P_{S',j} = \{z[j] \mid z \in \bigcup_{i=1}^n x_i'\}$ and 
$z[j]$ is $j$-th block of $z$. That is, $z[j]$ can be rewritten as $\Phi(\tilde{x}_j)$ for some $\tilde{x}_j$.
Note that, 
because $z$ is based on $\eta_{(x,y)}$ as shown in~\eqref{align:dk_z_ker},
$z[j]$ is in the Hilbert space $\Hilbert$ in the original problem.
Based on the relationship between $W'=(w'_1, \ldots, w'_k)$ and 
$W = (w_1, \ldots, w_k)$,
therefore, the hypothesis $h(x)$ in the original problem is
obtained by:
\begin{align}
    h(x) = &\arg\max_{j \in [k]} \langle w_j, \Phi(x) \rangle \\
    =&\arg\max_{j \in [k]} \langle w'_j, \Phi(x) \rangle \\
    =&\arg\max_{j \in [k]} \sum_{z[j] \in P_{S',j}} 
    \mu_{z[j]} \langle z[j], \Phi(x) \rangle \\
    =&\arg\max_{j \in [k]} \sum_{\tilde{x_j}} 
    \mu_{\tilde{x}_j} K(\tilde{x_j}, x).
\end{align}

\subsection{Reduction of other kernelized learning problems}
We can show that the other learning problems presented in this paper
can be kernelized.
For the other learning problems introduced in this study,
there are two types of the domains of $z$: the concatenation of the Hilbert vector
(complementarily labeled learning problems, multi-label learning, multi-task learning) and difference of the Hilbert vector (top-1 ranking learning).
For the difference in the Hilbert vector, that is, for $z = \Phi(x_1) - \Phi(x_2)$ and $\Phi(x)$,
$\langle z, \Phi(x) \rangle$ can be computed as: 
\begin{align}
    &\langle z, \Phi(x) \rangle\\
    =&\langle \Phi(x_1) - \Phi(x_2), \Phi(x) \rangle\\
    =& K(x_1, x) - K(x_2, x),
\end{align}
and thus $h(x)$ is computed by $h'$ in polynomial time.

\section{Comparison to the existing generalization bound for complementarily labeled learning}
\label{sec:ishida-comparison}
\citet{ishida2017learning} stated that, for a linear-hypothesis class,
the following bound holds with a probability of at least $1-\delta$:
$\RMC_{\calD}(h) \leq \widehat{R}(h) + ak(k-1) \frac{C_1C_2}{\sqrt{n}} + (k-1)\sqrt{\nicefrac{8 \ln (2/\delta)}{n}}$.
They used the empirical risk $\widehat{R}(h)$ for complementarily labeled instances, which is
different from the risk that we defined~\cite[see details in][]{ishida2017learning}.
According to this difference,
the proposed generalization bound is incomparable to the
existing bound.
However, we can say that if we achieve a small empirical risk
close to zero,
the proposed risk bound
is $k$ times tighter than the existing bound.


\section{Artificial datasets on complementarily labeled learning}
\label{sec:art_data}
We prepared three datasets, artificial1, artificial2, and artificial3.
Each dataset has 1000 training and 1000 test instances.
The number of dimension $d$ is 50.
They have 5, 10, and 25 classes, respectively.
The feature values of each data is determined by the following rule:
If the data belongs to class $j$, $\{\frac{(j-1)d}{k} +1, \ldots, \frac{jd}{k} \}$-th features
have the values drawn according to $\mathcal{N}(2, 1)$ and other features have the values drawn according to $\mathcal{N}(0, 1)$.






\bibliography{suehiro_124}

\begin{thebibliography}{33}
\providecommand{\natexlab}[1]{#1}
\providecommand{\url}[1]{\texttt{#1}}
\expandafter\ifx\csname urlstyle\endcsname\relax
  \providecommand{\doi}[1]{doi: #1}\else
  \providecommand{\doi}{doi: \begingroup \urlstyle{rm}\Url}\fi

\bibitem[Agarwal(2014)]{agarwal2014surrogate}
Shivani Agarwal.
\newblock Surrogate regret bounds for bipartite ranking via strongly proper
  losses.
\newblock \emph{The Journal of Machine Learning Research}, 15\penalty0
  (1):\penalty0 1653--1674, 2014.

\bibitem[Ailon and Mohri(2010)]{ailon2010preference}
Nir Ailon and Mehryar Mohri.
\newblock Preference-based learning to rank.
\newblock \emph{Machine Learning}, 80\penalty0 (2-3):\penalty0 189--211, 2010.

\bibitem[Andrews et~al.(2003)Andrews, Tsochantaridis, and
  Hofmann]{NIPS2002misvm}
Stuart Andrews, Ioannis Tsochantaridis, and Thomas Hofmann.
\newblock Support vector machines for multiple-instance learning.
\newblock In \emph{Advances in Neural Information Processing Systems}, pages
  577--584, 2003.

\bibitem[Balcan et~al.(2008)Balcan, Bansal, Beygelzimer, Coppersmith, Langford,
  and Sorkin]{balcan2008robust}
Maria-Florina Balcan, Nikhil Bansal, Alina Beygelzimer, Don Coppersmith, John
  Langford, and Gregory~B Sorkin.
\newblock Robust reductions from ranking to classification.
\newblock \emph{Machine learning}, 72\penalty0 (1-2):\penalty0 139--153, 2008.

\bibitem[Bartlett and Mendelson(2003)]{Bartlett:2003:RGC}
Peter~L. Bartlett and Shahar Mendelson.
\newblock Rademacher and gaussian complexities: Risk bounds and structural
  results.
\newblock \emph{Journal of Machine Learning Research}, 3:\penalty0 463--482,
  2003.

\bibitem[Beygelzimer et~al.(2005)Beygelzimer, Dani, Hayes, Langford, and
  Zadrozny]{beygelzimer2005error}
Alina Beygelzimer, Varsha Dani, Tom Hayes, John Langford, and Bianca Zadrozny.
\newblock Error limiting reductions between classification tasks.
\newblock In \emph{International Conference on machine Learning}, pages 49--56,
  2005.

\bibitem[Beygelzimer et~al.(2015)Beygelzimer, Daum{\'e}, Langford, and
  Mineiro]{beygelzimer2015learning}
Alina Beygelzimer, Hal Daum{\'e}, John Langford, and Paul Mineiro.
\newblock Learning reductions that really work.
\newblock \emph{Proceedings of the IEEE}, 104\penalty0 (1):\penalty0 136--147,
  2015.

\bibitem[Bhatia et~al.(2015)Bhatia, Jain, Kar, Varma, and
  Jain]{NIPS2015_35051070}
Kush Bhatia, Himanshu Jain, Purushottam Kar, Manik Varma, and Prateek Jain.
\newblock Sparse local embeddings for extreme multi-label classification.
\newblock In \emph{Advances in Neural Information Processing Systems},
  volume~28, 2015.

\bibitem[Boyd and Vandenberghe(2004)]{boyd-vandenberghe:book04}
Stephan Boyd and Lieven Vandenberghe.
\newblock \emph{Convex Optimization}.
\newblock Cambridge University Press, 2004.

\bibitem[Carbonneau et~al.(2018)Carbonneau, Cheplygina, Granger, and
  Gagnon]{CARBONNEAU2018329}
Marc-Andr\'{e} Carbonneau, Veronika Cheplygina, Eric Granger, and Ghyslain
  Gagnon.
\newblock Multiple instance learning: A survey of problem characteristics and
  applications.
\newblock \emph{Pattern Recognition}, 77:\penalty0 329 -- 353, 2018.

\bibitem[Dietterich et~al.(1997)Dietterich, Lathrop, and
  Lozano-P{\'e}rez]{Dietterich:1997}
Thomas~G. Dietterich, Richard~H. Lathrop, and Tom{\'a}s Lozano-P{\'e}rez.
\newblock Solving the multiple instance problem with axis-parallel rectangles.
\newblock \emph{Artificial Intelligence}, 89\penalty0 (1-2):\penalty0 31--71,
  1997.

\bibitem[Doran(2015)]{doran:thesis}
Gary Doran.
\newblock \emph{Multiple Instance Learning from Distributions}.
\newblock PhD thesis, Case WesternReserve University, 2015.

\bibitem[Doran and Ray(2014)]{Doran:2014}
Gary Doran and Soumya Ray.
\newblock A theoretical and empirical analysis of support vector machine
  methods for multiple-instance classification.
\newblock \emph{Machine Learning}, 97\penalty0 (1-2):\penalty0 79--102, 2014.

\bibitem[G\"{a}rtner et~al.(2002)G\"{a}rtner, Flach, Kowalczyk, and
  Smola]{Gartner02multi-instancekernels}
Thomas G\"{a}rtner, Peter~A. Flach, Adam Kowalczyk, and Alex~J. Smola.
\newblock Multi-instance kernels.
\newblock In \emph{International Conference on Machine Learning}, pages
  179--186, 2002.

\bibitem[Hidasi and Karatzoglou(2018)]{hidasi2018recurrent}
Bal{\'a}zs Hidasi and Alexandros Karatzoglou.
\newblock Recurrent neural networks with top-k gains for session-based
  recommendations.
\newblock In \emph{International Conference on Information and Knowledge
  Management}, pages 843--852, 2018.

\bibitem[Ishida et~al.(2017)Ishida, Niu, Hu, and Sugiyama]{ishida2017learning}
Takashi Ishida, Gang Niu, Weihua Hu, and Masashi Sugiyama.
\newblock Learning from complementary labels.
\newblock In \emph{Advances in neural information processing systems}, pages
  5639--5649, 2017.

\bibitem[Ishida et~al.(2019)Ishida, Niu, Menon, and Sugiyama]{ishida19a}
Takashi Ishida, Gang Niu, Aditya Menon, and Masashi Sugiyama.
\newblock Complementary-label learning for arbitrary losses and models.
\newblock In \emph{International Conference on Machine Learning}, volume~97,
  pages 2971--2980, 2019.

\bibitem[James and Hastie(1998)]{james1998error}
Gareth James and Trevor Hastie.
\newblock The error coding method and picts.
\newblock \emph{Journal of Computational and Graphical statistics}, 7\penalty0
  (3):\penalty0 377--387, 1998.

\bibitem[Kingma and Ba(2014)]{kingma2014adam}
Diederik~P Kingma and Jimmy Ba.
\newblock Adam: A method for stochastic optimization.
\newblock In \emph{International Conference on Learning Representations}, 2014.

\bibitem[Le~Thi and Dinh(2018)]{le2018dc}
Hoai~An Le~Thi and Tao~Pham Dinh.
\newblock Dc programming and dca: thirty years of developments.
\newblock \emph{Mathematical Programming}, 169\penalty0 (1):\penalty0 5--68,
  2018.

\bibitem[{Lei} et~al.(2019){Lei}, {Dogan}, {Zhou}, and {Kloft}]{Lei19}
Y.~{Lei}, \"{U}. {Dogan}, D.~{Zhou}, and M.~{Kloft}.
\newblock Data-dependent generalization bounds for multi-class classification.
\newblock \emph{IEEE Transactions on Information Theory}, 65\penalty0
  (5):\penalty0 2995--3021, 2019.

\bibitem[Mohri et~al.(2018)Mohri, Rostamizadeh, and
  Talwalkar]{mohri2018foundations}
Mehryar Mohri, Afshin Rostamizadeh, and Ameet Talwalkar.
\newblock \emph{Foundations of machine learning}.
\newblock MIT press, 2018.

\bibitem[Pitt and Warmuth(1990)]{pitt1990prediction}
Leonard Pitt and Manfred~K Warmuth.
\newblock Prediction-preserving reducibility.
\newblock \emph{Journal of Computer and System Sciences}, 41\penalty0
  (3):\penalty0 430--467, 1990.

\bibitem[Pontil and Maurer(2013)]{pontil2013excess}
Massimiliano Pontil and Andreas Maurer.
\newblock Excess risk bounds for multitask learning with trace norm
  regularization.
\newblock In \emph{Conference on Learning Theory}, pages 55--76. PMLR, 2013.

\bibitem[Ramaswamy et~al.(2014)Ramaswamy, Babu, Agarwal, and
  Williamson]{ramaswamy2014consistency}
Harish~G Ramaswamy, Balaji~Srinivasan Babu, Shivani Agarwal, and Robert~C
  Williamson.
\newblock On the consistency of output code based learning algorithms for
  multiclass learning problems.
\newblock In \emph{Conference on Learning Theory}, pages 885--902, 2014.

\bibitem[Sabato and Tishby(2012)]{Sabato:2012:MLA}
Sivan Sabato and Naftali Tishby.
\newblock Multi-instance learning with any hypothesis class.
\newblock \emph{Journal of Machine Learning Research}, 13\penalty0
  (1):\penalty0 2999--3039, 2012.

\bibitem[Suehiro et~al.(2020)Suehiro, Hatano, Takimoto, Yamamoto, Bannai, and
  Takeda]{suehiro2020multiple}
Daiki Suehiro, Kohei Hatano, Eiji Takimoto, Shuji Yamamoto, Kenichi Bannai, and
  Akiko Takeda.
\newblock Theory and algorithms for shapelet-based multiple-instance learning.
\newblock \emph{Neural Computation}, 32\penalty0 (8):\penalty0 1580--1613,
  2020.

\bibitem[Xu et~al.(2016{\natexlab{a}})Xu, Liu, Tao, and Xu]{xu2016local}
Chang Xu, Tongliang Liu, Dacheng Tao, and Chao Xu.
\newblock Local rademacher complexity for multi-label learning.
\newblock \emph{IEEE Transactions on Image Processing}, 25\penalty0
  (3):\penalty0 1495--1507, 2016{\natexlab{a}}.

\bibitem[Xu et~al.(2016{\natexlab{b}})Xu, Tao, and Xu]{xu2016robust}
Chang Xu, Dacheng Tao, and Chao Xu.
\newblock Robust extreme multi-label learning.
\newblock In \emph{ACM SIGKDD international conference on knowledge discovery
  and data mining}, pages 1275--1284, 2016{\natexlab{b}}.

\bibitem[Yu et~al.(2014)Yu, Jain, Kar, and Dhillon]{pmlr-v32-yu14}
Hsiang-Fu Yu, Prateek Jain, Purushottam Kar, and Inderjit Dhillon.
\newblock Large-scale multi-label learning with missing labels.
\newblock In \emph{International Conference on Machine Learning}, pages
  593--601, 2014.

\bibitem[Zhang et~al.(2013)Zhang, He, Si, and Lawrence]{pmlr-v28-zhang13a}
Dan Zhang, Jingrui He, Luo Si, and Richard Lawrence.
\newblock {MILEAGE}: Multiple instance learning with global embedding.
\newblock In \emph{International Conference on Machine Learning}, pages 82--90,
  2013.

\bibitem[Zhang et~al.(2020)Zhang, Liu, and Li]{Zhang2020RobustML}
Weijia Zhang, Lin Liu, and Jiuyong Li.
\newblock Robust multi-instance learning with stable instances.
\newblock In \emph{European Conference on Artificial Intelligence}, 2020.

\bibitem[Zhou and Xu(2007)]{zhou2007relation}
Zhi-Hua Zhou and Jun-Ming Xu.
\newblock On the relation between multi-instance learning and semi-supervised
  learning.
\newblock In \emph{International Conference on Machine Learning}, pages
  1167--1174, 2007.

\end{thebibliography}

\end{document}